\documentclass[11pt, a4paper, logo, copyright, nonumbering]{map}
\usepackage{CJKutf8}
\usepackage{xargs}  
\usepackage{todonotes}
\usepackage{amsmath}
\usepackage{dsfont}

\usepackage{longtable}

\usepackage{svg}
\usepackage{fontawesome}
\usepackage{array}
\usepackage{tabularx}
\usepackage{latexsym}
\usepackage{graphicx}
\usepackage[utf8]{inputenc} 
\usepackage[utf8]{inputenc} 
\usepackage{amssymb} 
\usepackage{url}            
\usepackage{amsfonts}       
\usepackage{nicefrac}       
\usepackage{microtype}      
\usepackage{xspace}

\usepackage{listings}
\usepackage{makecell}
\usepackage{inconsolata}
\usepackage{multicol}
\usepackage{amstext}

\definecolor{darkblue}{RGB}{84, 112, 198}

\definecolor{lightblue}{rgb}{0.85, 0.95, 1.0}    
\definecolor{lightgreen}{rgb}{0.90, 1.0, 0.90}    
\definecolor{lightorange}{rgb}{1.0, 0.95, 0.85}   
\definecolor{lightpurple}{rgb}{0.95, 0.90, 1.0}   
\definecolor{lightgray}{rgb}{0.97, 0.97, 0.97}    

\usepackage{tikz}
\usetikzlibrary{calc}
\definecolor{battery-empty}{rgb}{0.9, 0.9, 0.9}
\newcommand{\difficultybar}[1]{%
  \begin{tikzpicture}[baseline, scale=0.5, every node/.style={scale=0.8}]
    \foreach \i in {1,2,3,4,5} {
      \ifnum\i>#1
        \draw[fill=battery-empty] (\i*0.5-0.5, 0) rectangle (\i*0.5, 0.25);
      \else
        \pgfmathsetmacro{\colorlevel}{80 - 12*(\i)} 
        \edef\x{\noexpand\draw[fill=blue!\colorlevel!white, opacity=0.9] (\i*0.5-0.5, 0) rectangle (\i*0.5, 0.25);}
        \x
        \draw[blue!50!black] (\i*0.5-0.5, 0) rectangle (\i*0.5, 0.25);
      \fi
    }
    \fill[battery-empty!70] (2.5, 0.08) rectangle (2.6, 0.17);
    \draw[battery-empty!70!black] (2.5, 0.08) rectangle (2.6, 0.17);
  \end{tikzpicture}%
}
\renewcommand{\arraystretch}{0.96}

\definecolor{hidden-draw}{RGB}{20,68,106}
\definecolor{hidden-pink}{RGB}{255,245,247}
\usepackage[edges]{forest}

\usepackage{bm}

\usepackage{natbib}
\usepackage{CJKutf8}
\usepackage{xargs}
\usepackage{todonotes}
\usepackage{multirow}
\usepackage{cleveref}
\usepackage{subcaption}
\usepackage{url}
\usepackage{svg}
\usepackage{fontawesome}
\usepackage{xcolor}
\usepackage{float}

\usepackage[utf8]{inputenc}
\usepackage{booktabs}
\usepackage{graphicx}
\usepackage{array}
\usepackage{tabularx}
\usepackage{xcolor,colortbl}  
\definecolor{boxcolor}{HTML}{d92523} 
\definecolor{bulbcolor}{HTML}{e3b87f} 
\usepackage{url}
\usepackage{ragged2e}   
\usepackage{makecell} 

\usepackage{hyperref}       
\usepackage{url}            
\usepackage{booktabs}       
\usepackage{nicefrac}       
\usepackage{microtype}      
\usepackage{xcolor}         
\usepackage{graphicx}
\usepackage{longtable}
\usepackage{array}
\usepackage{pbox}
\usepackage{tabularx}
\usepackage{multirow}
\usepackage{wrapfig}
\usepackage{pifont}
\usepackage{lipsum}
\usepackage{subcaption}
\usepackage{enumitem}
\usepackage{tikz}
\usepackage{xstring}

\usepackage{color-edits}

\usepackage{booktabs}
\usepackage{multirow}
\usepackage[table]{xcolor}
\usepackage{tabularx}

\usepackage{parskip} 

\usepackage{threeparttable} 

\usepackage{algorithm}
\usepackage{algpseudocode}

\usepackage{xspace}
\usepackage{times}
\usepackage{latexsym}
\usepackage{booktabs}
\usepackage{fvextra}
\usepackage{graphicx}
\usepackage{subcaption} 
\usepackage[T1]{fontenc}
\usepackage{xcolor}
\usepackage[utf8]{inputenc}

\usepackage{microtype}

\usepackage{inconsolata}
\usepackage{amsmath}
\usepackage{booktabs}
\usepackage{graphicx}
\usepackage{pifont}
\usepackage{multirow}          
\usepackage{colortbl}          
\usepackage{algorithm}
\usepackage{algpseudocode}
\usepackage{xcolor}
\usepackage{amssymb}

\definecolor{rliableolive}{HTML}{BBCC33}
\definecolor{rliableblue}{HTML}{77AADD}
\definecolor{rliablered}{HTML}{f63c44}

\algrenewcommand\algorithmicrequire{\textbf{Input:}}
\algrenewcommand\algorithmicensure{\textbf{Output:}}
\algrenewcommand\alglinenumber[1]{\small #1:}

\newcommand{\ie}{\textit{i.e.,}\xspace}

\newcommand{\paratitle}[1]{\vspace{1.5ex}\noindent\textbf{#1}}

\usepackage{alltt}
\usepackage[most,skins,theorems]{tcolorbox}
\tcbuselibrary{skins,breakable,fitting,hooks}  
\tcbset{
  aibox/.style={
    width=\linewidth,
    top=7pt,
    bottom=8pt,
    colback=white,
    colframe=black,
    colbacktitle=black,
    enhanced,
    center,
    attach boxed title to top left={yshift=-0.1in,xshift=0.15in},
    boxed title style={boxrule=0pt,colframe=white},
  }
}
\tcbset{
  aibox2/.style={
    width=\linewidth,
    top=7pt,
    bottom=2pt,
    colback=green!18!white,
    colframe=black,
    colbacktitle=black,
    enhanced,
    center,
    attach boxed title to top left={yshift=-0.1in,xshift=0.15in},
    boxed title style={boxrule=0pt,colframe=white,},
  }
}
\newtcolorbox{AIbox}[2][]{aibox,title=#2,#1}
\newtcolorbox{AIbox2}[2][]{aibox2,title=#2,#1}

\usepackage{xcolor}

\newcommand{\goodhl}[1]{\colorbox{green!18}{\strut #1}}
\newcommand{\badhl}[1]{\colorbox{red!18}{\strut #1}}
\hypersetup{
    colorlinks=true,            
    linkcolor=blue,             
    filecolor=magenta,          
    urlcolor=cyan,              
    citecolor=purple,             
    pdftitle={Overleaf Example},
    pdfpagemode=FullScreen,
}

\definecolor{iquestblue}{HTML}{173C7F}
\definecolor{iquestazure}{HTML}{528FCC}

\newcommandx{\info}[2][1=]{\todo[linecolor=red,backgroundcolor=red!25,bordercolor=red,#1]{#2}}

\title{
\vspace{-0.2in}
\centering \fontsize{15pt}{16pt}\selectfont
ClawGym: A Scalable Framework for Building Effective Claw Agents

\vspace{-0.2in}
}

\author{
Fei Bai\textsuperscript{1*},
Huatong Song\textsuperscript{1*},
Shuang Sun\textsuperscript{1*},
Daixuan Cheng\textsuperscript{1}, 
Yike Yang\textsuperscript{1},
Chuan Hao\textsuperscript{2},
Renyuan Li\textsuperscript{2},\\
\normalfont
Feng Chang\textsuperscript{2},
Yuan Wei\textsuperscript{2},
Ran Tao\textsuperscript{2},
Bryan Dai\textsuperscript{2},
Jian Yang\textsuperscript{3},
Wayne Xin Zhao\textsuperscript{1$\dagger$},
Ji-Rong Wen\textsuperscript{1}\\
\normalfont
\textsuperscript{1}Gaoling School of Artificial Intelligence, Renmin University of China, \\
\normalfont
\textsuperscript{2}IQuest Research,
\textsuperscript{3}Beihang University \\
\normalfont
\normalsize{\textsuperscript{*}Equal contributors alphabetically ordered \quad
\textsuperscript{$\dagger$}Corresponding Author \\
\normalfont
Email: \texttt{\{feibai,songhuatong123,sunshuang\}@ruc.edu.cn}, \\
\normalfont
\texttt{batmanfly@gmail.com}
}}

\newcommand{\ignore}[1]{}

\usepackage[most]{tcolorbox}
\definecolor{darkorange}{RGB}{255, 140, 0}
\definecolor{lightgreen}{RGB}{145, 204, 117}
\definecolor{lightyellow}{RGB}{250, 200, 88}
\definecolor{lightred}{RGB}{238, 102, 102}
\definecolor{lightblue}{RGB}{115, 192, 222}

\definecolor{gray_1}{HTML}{B7B7B7}
\definecolor{gray_2}{HTML}{F0F0F0} 
\definecolor{frame_blue}{HTML}{A9D18E}

\newtcolorbox[auto counter, number within=section]{PromptBox}[2][]{
    enhanced,
    breakable,
    colback=gray_2, 
    colframe=gray_1,
    coltitle=white,
    fontupper=\small,
    fonttitle=\bfseries,
    title={#2}, 
    label={#1},
    arc=2pt,
    boxrule=1pt,
    left=2mm, right=2mm, top=2mm, bottom=2mm,
}

\usepackage{etoolbox}
\makeatletter
\patchcmd{\@maketitle}
  {\@toptitlebar}
  {%
    \hbox to \textwidth{%
    
    \includegraphics[height=0.7cm]{figures/aibox-logo.png}%
      \hfill
      \includegraphics[height=0.7cm]{figures/logo2.pdf}%
    }%
    \vspace{0.3em}
    \@toptitlebar
  }
  {}{}
\makeatother

\begin{abstract}
Claw-style environments support multi-step workflows over local files, tools, and persistent workspace states. However, scalable development around these environments remains constrained by the absence of a systematic framework, especially one for synthesizing verifiable training data and integrating it with agent training and diagnostic evaluation.
To address this challenge, we present \textbf{ClawGym}, a scalable framework that supports the full lifecycle of Claw-style personal agent development.
Concretely, we construct \textbf{ClawGym-SynData}, a diverse dataset of 13.5K filtered tasks synthesized from persona-driven intents and skill-grounded operations, paired with realistic mock workspaces and hybrid verification mechanisms. 
We then train a family of capable Claw-style models, termed \textbf{ClawGym-Agents}, through supervised fine-tuning on black-box rollout trajectories, and further explore reinforcement learning via a lightweight pipeline that parallelizes rollouts across per-task sandboxes.
To support reliable evaluation, we further construct \textbf{ClawGym-Bench}, a benchmark of 200 instances calibrated through automated filtering and human-LLM review.
Relevant resources have been released at \url{https://github.com/ClawGym}.

\end{abstract}

\begin{document}

\maketitle

\let\oldthefootnote\thefootnote



\section{Introduction}
\label{sec-intro}

Recent advancements in autonomous agent frameworks, such as OpenClaw~\citep{openclaw2026}, have fundamentally reshaped how AI integrates with daily digital life and are increasingly used to tackle real-world tasks. Unlike traditional chatbots confined to isolated dialog boxes~\citep{zhu2025evolutionary}, these systems are deployed directly within users' computer environments~\citep{cheng2026computerenvironmentselicitgeneral}, where they can invoke tools, manage local file systems, and interact with rich web-enabled services.
This paradigm bridges the gap between machine intelligence and digital execution, sparking a new wave of applications in which users ``raise'' digital agents to perform concrete, multi-step tasks in their real-world daily workflows~\citep{ma2026skillclaw}.

Despite the impressive capabilities of OpenClaw and similar systems~\citep{qwenpaw2026, nanobot2026}, autonomous agents still struggle to reliably handle many everyday digital tasks, even when powered by the most advanced large language models (LLMs)~\citep{zhao2023survey}. 
This limitation becomes even more pronounced for less capable LLMs, which often fail to follow multi-step instructions, select appropriate tools, or recover from execution errors in the OpenClaw environment.
The gap reflects the distinctive nature of Claw-style tasks.
Unlike static text-based reasoning tasks such as AIME~\citep{aime2026}, which are well specified and solvable through pure reasoning~\citep{bai2025effectivecodeintegratedreasoning}, Claw-style tasks are grounded in local workspace states, requiring agents to reason over existing artifacts, execute tools, and update the workspace through multi-step interactions.
They also differ from conventional agentic benchmarks such as SWE-Bench-Verified~\citep{sweb-verified} and BrowseComp~\citep{wei2025browsecomp}, which typically provide more structured and observable agent loops or environments~\citep{sun2025simpledeepsearcherdeepinformationseeking, arpo, bai2025effectivecodeintegratedreasoning}.
In Claw-style settings, agents instead operate through opaque system interfaces, where they must handle ambiguous instructions, unexpected workspace states, tool execution errors, and long-horizon dependencies across sessions.
Accordingly, these properties call for workspace-grounded evaluation beyond static reasoning and structured agent-loop benchmarks.

Recognizing these challenges, recent studies have sought to improve personal agents through either specialized training algorithms~\citep{clawr1-2026, xia2026metaclaw} or evaluation benchmarks~\citep{ye2026claw, Ding_WildClawBench}.
On the training side, specialized algorithms~\citep{clawr1-2026, xia2026metaclaw} and online learning frameworks~\citep{wang2026openclawrl} have been proposed to extract supervision or feedback signals from daily user interactions, enabling agents to improve through continued use. 
On the evaluation side, dedicated benchmarks~\citep{ye2026claw, Ding_WildClawBench, meng2026clawmarklivingworldbenchmarkmultiturn, pinchbench2026} have been constructed across diverse daily tasks, providing valuable resources for assessing personal agent capabilities.

Nevertheless, existing studies are often limited by the lack of large-scale Claw-style task data grounded in users' workspaces. Compared with static text-based tasks and structured agent benchmarks~\citep{aime2026, sweb-verified, wei2025browsecomp}, Claw-style agent tasks introduce distinct challenges for large-scale data synthesis, driven by three key factors.
First, they need to capture personalized requirements across different professions and routines, making it difficult to define sufficiently representative task settings that cover a broad range of realistic scenarios~\citep{patwardhan2025gdpvalevaluatingaimodel}. Second, their long-horizon nature often involves sequences of file operations, tool calls, workspace updates, and intermediate validation, which poses substantial verifiability challenges for automated assessment~\citep{sun2025scaling}. Third, their grounding in local workspaces necessitates realistic mock workspaces and task-specific artifacts that provide meaningful execution contexts~\citep{xie2024osworld, bonatti2024windows}. Given the limited availability of Claw-style task data, it remains underexplored how to develop systematic frameworks that support both agent training in Claw-style environments and broader benchmark construction. This gap hinders progress towards more capable environment-grounded agent systems.

In light of this challenge, we propose \textbf{ClawGym}, a data-centric framework designed to systematically unify task synthesis, agent training, and performance evaluation for developing Claw-style personal agents. At the core of our approach lies a dual-route data synthesis strategy that generates both diverse and verifiable tasks. Specifically, the persona-driven top-down pipeline grounds tasks in high-level user personas and scenario intents, while the skill-grounded bottom-up pipeline composes multi-step workflows from concrete, executable capabilities. To ensure execution realism, we automatically instantiate workspace-grounded environments by generating task-specific mock files and auxiliary resources. We further implement a hybrid verification protocol that combines deterministic code-based checkers with qualitative rubric-based verifiers, providing robust judgment for each task. 
Following this dual-route approach, we synthesize and curate 13.5K Claw-style tasks within the OpenClaw framework. 
Collectively, these tasks constitute \textbf{ClawGym-SynData}, a large-scale synthesized dataset for training Claw-style agents.


The availability of such data further enables the design of suitable agent training algorithms. Building on this synthesized task pool, we develop \textbf{ClawGym-Agents}, a family of agent models trained on Claw-style task data by collecting and filtering high-fidelity interaction trajectories through large-scale black-box rollouts on the OpenClaw harness, followed by supervised fine-tuning (SFT) to improve LLM performance within the OpenClaw framework. Furthermore, we explore reinforcement learning (RL) through a lightweight sandbox-parallel pipeline.





To assess the effectiveness of our approach, we construct a reliable evaluation benchmark, termed \textbf{ClawGym-Bench}, by selecting discriminative tasks from ClawGym-SynData (excluding those used in the training set). This selection process adopts a rigorous pipeline that combines rollout-based difficulty calibration with LLM-assisted human review. Only samples that pass both automated assessments and human review are retained as benchmark instances, establishing a robust diagnostic platform for evaluating agent performance in Claw-style scenarios. 
When evaluating our approach on ClawGym-Bench alongside another public benchmark, PinchBench, we empirically find that smaller LLMs trained on our synthesized data achieve substantial improvements in performing OpenClaw tasks: Qwen3-8B improves by 38.90\% on PinchBench and 43.46\% on ClawGym-Bench. Larger LLMs also show clear gains, with Qwen3-30B-A3B improving by 54.68\% on PinchBench and 25.96\% on ClawGym-Bench. Beyond aggregate performance, we further conduct behavioral analyses of Claw-style agents, offering insights for future research on robust environment-grounded agent development.

In summary, we propose ClawGym, a scalable framework that integrates data synthesis, agent training, and performance evaluation for building effective Claw agents.
 Our work aims to advance research on Claw-style agents.
The contributions of ClawGym are threefold:





\begin{itemize}[leftmargin=*]
\item \textbf{ClawGym-SynData}: 
We release the first large-scale synthesized dataset for Claw agents, containing 13.5K executable tasks. Its automated synthesis pipeline combines persona-driven intents with skill-grounded operations, enabling scalable generation of diverse, verifiable, and claw-style training data.

\item \textbf{ClawGym-Agents}: We collect high-quality trajectories via black-box rollouts on OpenClaw and perform supervised fine-tuning to obtain a family of capable Claw agents; we further explore reinforcement learning via a lightweight RL pipeline with sandbox parallelism, delivering consistent gains.

\item \textbf{ClawGym-Bench}: 
We construct a benchmark of 200 instances across six categories for evaluating Claw agents, with reliability ensured through difficulty-aware filtering and human-LLM review. The benchmark reveals clear capability gaps among Claw agents across different model scales. 
\end{itemize}

\section{Preliminary}
In this section, we first present the task definition and then briefly introduce the framework overview.

\subsection{Task Definition}

In our work, a Claw-style  agent  task is an environment-grounded instruction-execution problem. The agent is provided with a user instruction and an initial workspace, and it may utilize available computer tools within that workspace to perform work in an effort to satisfy the user's request. Formally, a task instance is denoted as:  
\begin{equation}
    \tau = \langle  p, s_0, \mathcal{A}, \mathcal{F}, \mathcal{V}_\tau\rangle,
\end{equation}
where $p$ is the user instruction, $\mathcal{A}$ is the set of actions available to the agent, $\mathcal{F}$ describes how tool execution updates the environment state, and $\mathcal{V}_\tau$ is the task-specific verifier.

\paragraph{Input.}
The input to the agent consists of two parts: the user instruction $p$ and the initial environment state $s_0$. 
The instruction $p$ describes the user's goal, such as organizing files, extracting information, editing documents, generating reports, or configuring software. 
The environment state $s_0$ provides the concrete workspace  for the task. 
It may include local files and folders, web-accessible interfaces, configuration files, and other workspace artifacts. 
The agent can interact with the environment through actions from the available action space $\mathcal{A}$. 
Each action corresponds to a tool call or executable operation, such as reading a file, writing an output file, moving or deleting files, running a script, querying the system, or interacting with a web page. 
Together, $p$, $s_0$, and $\mathcal{A}$ specify both what the agent needs to accomplish and what operations it can use.

\paragraph{Output.}
To solve the task, the agent produces a trajectory composed of
action and observation segments:
\begin{equation}
    \xi = (A_1, O_1, A_2, O_2, \ldots, A_K, O_K),
\end{equation}
where each action segment
$A_k=(a_{k,1},\ldots,a_{k,m_k})$ contains one or more consecutive executable
actions with $a_{k,i}\in\mathcal{A}$, and each observation segment
$O_k=(o_{k,1},\ldots,o_{k,n_k})$ contains one or more environment observations.
This segment-level formulation captures non-alternating interaction patterns in
Claw-style execution, where an agent may issue multiple tool calls before
receiving or processing the corresponding feedback. Standard strict
action-observation alternation is a special case where $m_k=n_k=1$ for all $k$.

Let $(a_1,\ldots,a_H)$ denote the flattened sequence of executable actions in
$\xi$, where $H=\sum_{k=1}^{K}m_k$. After each executable action, the workspace
state may update as
\begin{equation}
    s_t = \mathcal{F}(s_{t-1}, a_t), \quad t = 1,\ldots,H.
\end{equation}
After $H$ executable actions, the agent reaches a final environment state $s_H$.
The agent may also return a natural language response $y$, such as a completion
confirmation, a progress summary, or an explanation of failure. However, for
Claw-style agent tasks, the primary output is the final environment state $s_H$,
rather than merely the final text response. This final state may contain newly
created files, modified documents, reorganized directories, generated reports, or
other persistent workspace changes.

\paragraph{Verification Criteria.} 
A task is considered successfully completed if the final environment state satisfies the user's original instruction. 
Given the initial state $s_0$, the final state $s_H$, and the optional final response $y$, the task-specific verifier assigns a score:
\begin{equation}
    v = \mathcal{V}_\tau(s_0, s_H, y), \quad v \in [0,1].
\end{equation}
The task receives full credit when $v=1$.
In principle, a complete evaluation may also consider whether the agent's intermediate actions are safe, efficient, and robust to errors. 
However, such trajectory-level properties are difficult to verify consistently for open-ended environment-based tasks, because different agents may complete the same task through different valid action sequences. 
Therefore, this work primarily focuses on final-state correctness, while leaving more detailed assessment of action safety, efficiency, and error recovery to future work. 
The verifier $\mathcal{V}_\tau$ can be  instantiated using  code-based, rubric-based verification, or  a combination of both~\citep{qwenclawbench1.1}. 
Code-based checkers examine deterministic workspace properties, such as whether required files exist, whether output schemas are valid, whether computed values are correct, and whether generated artifacts satisfy task-specific constraints. 
Rubric-based verification complements these checks by assessing qualitative aspects that are harder to capture with executable code, such as clarity, completeness, usefulness, faithfulness, and alignment with user intent.

\paragraph{Example.}
Consider the instruction: ``\textit{Find all CSV files in the project folder whose filenames contain the word `sales', merge their rows into a single CSV file, add a column named `source\_file' indicating which file each row comes from, and save the result as `sales\_report.csv' in the output folder.}'' 
Here, $p$ is the user-specified task prompt, $s_0$ contains the original project folder and available files, and $\mathcal{A}$ includes file-system operations and CSV-processing tools. 
The agent must execute a trajectory $\xi$ that identifies the relevant CSV files, reads their contents, validates their compatible schemas, merges the rows, records the source filename for each row, and saves the final file to the required location. 
The task is successfully completed if $\mathcal{V}_\tau$ confirms the final state $s_H$ contains a correct `\textit{sales\_report.csv}' file at the specified path.

\begin{figure}[t]
    \centering
    \includegraphics[width=1.0\linewidth]{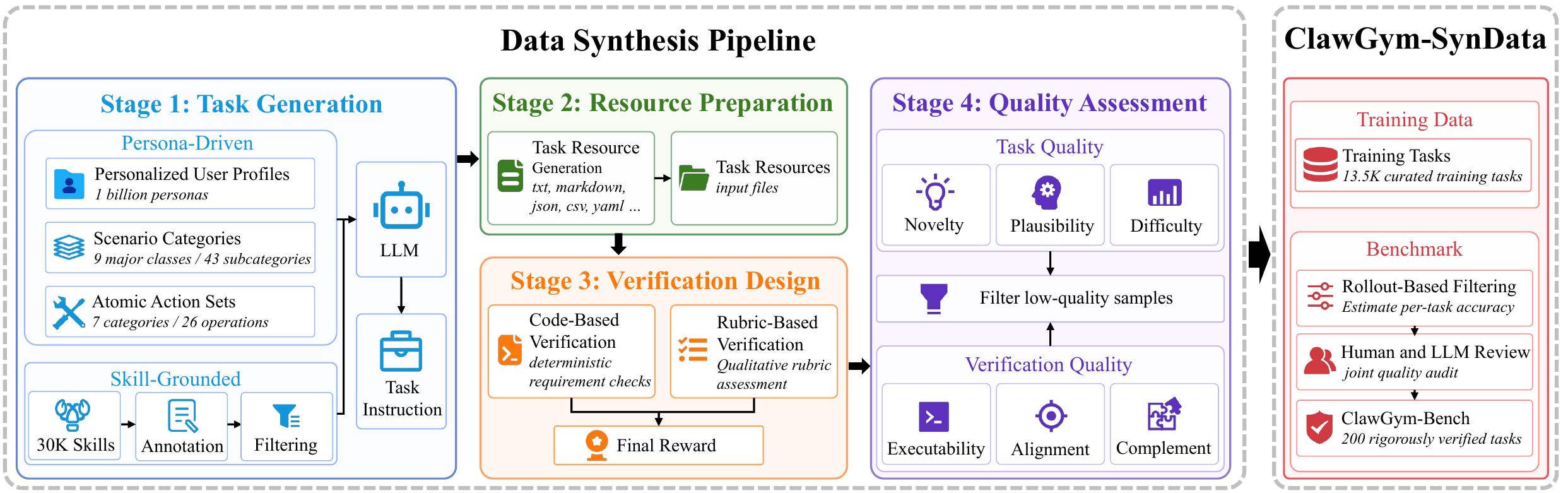}
    \caption{Overview of the ClawGym-SynData pipeline, which generates tasks from persona-driven and skill-grounded sources, prepares task resources, designs hybrid verification, filters samples through quality assessment, and constructs training and benchmark data.}
    \label{fig:main}
\end{figure}

\subsection{Framework Overview}
The \textbf{ClawGym} framework comprises three main components. First, \textbf{ClawGym-SynData} provides a carefully constructed synthesized dataset of 13.5K training tasks, with an automated synthesis process that enables large-scale data generation. Second, \textbf{ClawGym-Agents} trains a family of Claw agents on high-quality trajectories collected from synthesized tasks through black-box rollouts using the OpenClaw harness. Third, \textbf{ClawGym-Bench} offers a benchmark of 200 task instances for evaluating Claw-style agents, with carefully assessed reliability. Together, these components provide a foundation for advancing research on environment-grounded instruction-execution agents.
In what follows, we introduce ClawGym-SynData, ClawGym-Agents, and ClawGym-Bench in Sections~\ref{sec:data}, \ref{sec:agents}, and \ref{sec:bench}, respectively. 

\section{ClawGym-SynData: Scaling Effective Task Synthesis for Claw Agents}
\label{sec:data}

Our data synthesis framework is designed to generate scalable, high-quality datasets for Claw agents, addressing the lack of effective training data.
 The synthesis process consists of four main stages: 1) \textbf{Task Generation}, which follows two synergistic pipelines: \emph{persona-driven top-down synthesis}, producing tasks grounded in diverse user scenarios, and \emph{skill-grounded bottom-up synthesis}, constructing tasks by combining individual OpenClaw skills~\citep{clawhub2026} into realistic, multi-step workflows; 2) \textbf{Resource Preparation}, which creates lightweight auxiliary files and workspace artifacts needed to execute the tasks; 3) \textbf{Verification Design}, generating code-based and rubric-based checks; and 4) \textbf{Automated Quality Assessment}, evaluating task and verification quality. By combining these stages, ClawGym-SynData achieves both scenario diversity and execution realism, enabling the creation of datasets at a scale sufficient for large-scale training and benchmark construction while maintaining verifiable quality.
 
\subsection{Task Generation}

\subsubsection{Persona-Driven Top-Down Synthesis}

Persona-driven top-down synthesis generates diverse and practical Claw agent tasks from high-level user contexts. The pipeline first creates task seeds by synthesizing personalized user profiles, scenario categories, and atomic action sets. These task seeds specify who the user is, what type of scenario the task belongs to, and what operations the agent may need to perform. An LLM then expands each task seed into a concrete, multi-step task. Here, ``top-down'' indicates that the task is gradually contextualized by first generating user contexts, then instantiated and enriched with details. This design emphasizes practical needs and scenario diversity while keeping the generated tasks grounded in feasible agent operations.

\paragraph{Seed Formulation.}
In the first step, we construct a task seed that determines the high-level context and operational scope of a synthesized task. 
Specifically, each seed is formed by combining a user persona \(u \in \mathcal{U}\), a scenario category \(c \in \mathcal{C}\), and a set of basic operations \(\mathcal{G} = \{g_1, g_2, \dots, g_n\}\):
\begin{equation}
z = (u, c, \mathcal{G}).
\label{eq:task_seed}
\end{equation}

Here, \(u\) specifies the intended user background, including the user's goals, preferences, and working context; \(c\) specifies the scenario type of the task; and \(\mathcal{G}\) specifies the basic operations that the generated instruction should involve. 
Personas are sampled from a diverse collection of user profiles, covering different occupations, workflows, preferences, and daily needs. 
The scenario taxonomy contains 9 major classes and 43 subcategories, from which one subcategory is selected for each seed. 
The atomic-operation taxonomy contains 7 categories and 26 distinct operations, covering representative Claw-style activities such as web information retrieval, file processing, document editing, data analysis, script execution, and result reporting. 
By combining these three sources, the pipeline produces task seeds that are both diverse and operationally grounded. 
The persona and scenario category encourage realistic user-facing task intents, while the selected basic operations constrain the later generated instruction to actions that can plausibly be carried out in a Claw-style workspace. 

\paragraph{Seed-Guided Task Instruction Generation.}
In the second step, we instantiate a task-generation prompt template with the task seed \(z\). 
Given \(z=(u,c,\mathcal{G})\), the prompt template \(\pi(\cdot)\) is used to condition a task generator \(\mathcal{M}_\mathrm{task}\) (\ie GPT-5) to generate the user-facing task instruction:
\begin{equation}
p = \mathcal{M}_\mathrm{task}(\pi(z)),
\label{eq:task_generation}
\end{equation}
where \(p\) corresponds to the instruction component in the task definition, specifying the user's goal, context, expected operations, and output requirements. 
This formulation reflects the top-down nature of the pipeline: the persona \(u\) and scenario category \(c\) determine the task direction, while the selected basic operations \(\mathcal{G}\) constrain the instruction to plausible workspace operations. 
Different combinations of personas, categories, and operation sets therefore produce diverse concrete user requests, which capture realistic daily workflows and are later paired with \(s_0\) and \(\mathcal{V}_{\tau}\) to form complete Claw-style task instances.

\subsubsection{Skill-Grounded Bottom-Up Synthesis}

Unlike the persona-driven top-down pipeline, which starts from user contexts and scenario intentions, the skill-grounded bottom-up pipeline starts from concrete capabilities already implemented in OpenClaw skills. We collect raw skills from ClawHub~\citep{clawhub2026}, where each skill is typically organized as a directory containing files such as \texttt{SKILL.md}, \texttt{README.md}, or other usage descriptions. Each skill is treated not as a complete task, but as a reusable capability unit for constructing environment-grounded workflows. The purpose of this pipeline is to identify skills suitable for task synthesis, normalize capability descriptions, and compose them into practical task specifications. In this way, synthesized tasks are grounded in OpenClaw operations while  allowing diverse user needs and scenarios to emerge from skill composition.

\paragraph{Skill Annotation and Filtering.}
Given a raw skill collection \(\mathcal{K}=\{k_i\}_{i=1}^{N}\), we first use a powerful LLM (\ie MiniMax-M2.5) to convert each skill into a structured annotation with a skill annotator $\mathcal{M}_{\mathrm{ann}}$:
$\alpha_i = \mathcal{M}_{\mathrm{ann}}(\rho(k_i))
\label{eq:skill_annotation}$,
where \(\rho(\cdot)\) is the annotation prompt, and \(\alpha_i\) summarizes the key information needed for task synthesis, including the skill's summary, core content, usage constraints, input-output characteristics, and a binary synthesizability label \(y_i \in \{0,1\}\). 
The label \(y_i=1\) indicates that the skill is suitable for reliable task synthesis under our setting, while \(y_i=0\) indicates that it should be filtered out, for example because it depends on unavailable credentials, lacks sufficient task-relevant details, or is incompatible with the target task format. 
We then retain the synthesizable skills as:
\begin{equation}
\mathcal{K}^{+}
=
\{(k_i,\alpha_i) \mid k_i \in \mathcal{K},\ y_i=1\}.
\label{eq:skill_filtering}
\end{equation}

This annotation serves two primary purposes: filtering out skills unsuitable for reliable task synthesis under our target setting and providing normalized capability descriptions for the task generator (\ie GPT-5). Specifically, we annotate approximately 30K raw OpenClaw skills and retain 16K synthesizable skills after filtering, which are then used for downstream task synthesis.

\paragraph{Skill-Composition Task Construction.}
After filtering, we construct each task from one primary skill and a set of randomly sampled supporting skills. The primary skill defines the central capability and main objective of the task, while the supporting skills introduce auxiliary operations or contextual requirements. In our implementation, each task uses one primary skill and up to three optional supporting skills. The selected skill bundle is provided to a generation model to produce a user-facing task instruction:
\begin{equation}
p
=
\mathcal{M}_{\mathrm{task}}
\big(
\pi(k_{\mathrm{main}}, \mathcal{K}_{\mathrm{support}})
\big),
\label{eq:skill_task_generation}
\end{equation}
where \(k_{\mathrm{main}}\) denotes the primary skill, \(\mathcal{K}_{\mathrm{support}}\) denotes the supporting skill set, and \(\pi(\cdot)\) is the task-construction prompt. In this process, the task generator (\ie GPT-5) can condition on the original skill content together with its annotation, so that the generated instruction remains grounded in concrete operational details, usage constraints, and input-output requirements. We also support using the annotated core content instead of the raw skill document, which abstracts the skill into a more general capability and helps introduce controlled task diversity. The generated instruction describes the user request, required resources, expected behaviors, and verification targets in a unified format. Since each task is constructed from one primary skill and optional supporting skills, the generated instruction can combine a clear main objective with auxiliary operations, encouraging multi-step coordination across capabilities while remaining grounded in executable OpenClaw operations.

\subsection{Resource Preparation}

Claw-style tasks often require concrete files to establish the initial workspace state. 
Given the synthesized instruction \(p\), resource preparation constructs the file-based component of \(s_{0}\). 
Instead of relying on real user files or large external datasets, we generate lightweight mock files tailored to each task, such as documents to summarize, tables to analyze, configuration files to inspect, records to reorganize, or reference materials that constrain the expected output.

For each instruction, we first identify its file requirements, including filenames, directory paths, formats, key fields, and content constraints. 
These requirements are represented as a resource specification:
\begin{equation}
f = \{(l_i, t_i, d_i)\}_{i=1}^{m},
\label{eq:resource_spec}
\end{equation}
where \(l_i\) denotes the file path, \(t_i\) denotes the file type, and \(d_i\) denotes the file content specification. We then use an LLM-based resource generator (\ie GPT-5) to materialize f into concrete files and place them at the specified workspace paths.

The generated files are designed to be task-specific and verifiable. 
For text or markdown files, the content provides the entities, constraints, and references needed by the instruction. 
For structured files such as JSON, CSV, or YAML, we generate explicit schemas, tables, fields, and values, so that later checkers can recompute required statistics or validate output consistency against the inputs. 
This controlled construction makes each task self-contained, reproducible, and consistent with its verifier, while avoiding privacy risks and uncertainty from real user files.

\subsection{Verification Design}

After a Claw agent executes a task, a central challenge is to determine whether the task has been completed correctly. 
Unlike text-only reasoning and conventional agentic benchmarks with binary reward distributions, Claw tasks are executed in a workspace-grounded environment, where success may depend on generated files, modified artifacts, data transformations, structured outputs, or final responses. 
These requirements are heterogeneous: some can be checked deterministically, while others require qualitative judgment. 
To address this, we design a hybrid verification scheme consisting of code-based verification and rubric-based verification~\citep{pinchbench2026}. 
Code-based verification targets objective and executable requirements, such as whether required files are created, whether structured outputs follow the specified schema, or whether computed values are correct. 
Rubric-based verification complements it by assessing qualitative requirements that are difficult to formalize as deterministic checks, such as  clarity, completeness and style. We use an LLM-based verifier generator (\ie GPT-5) to produce both the executable checkers and rubric rules for each synthesized task.

\paragraph{Code-Based Verification.}

For each synthesized task, we generate an executable checker to verify objective requirements. 
These requirements are decomposed into a set of atomic verification points:
\begin{equation}
\mathcal{C} = \{c_1, c_2, \dots, c_m\},
\end{equation}
where each \(c_i\) corresponds to a deterministic condition, such as the existence of an output file, the correctness of a JSON field, the validity of a computed statistic, or the presence of required content in a generated document. 
Since some task requirements are implicit in the provided files rather than explicitly stated in the user instruction, the checker also performs cross-validation between the input resources and the generated outputs. 
For example, it may recompute statistics from an input CSV, verify that an output report includes all eligible records from a source file, or ensure that generated artifacts are consistent with constraints defined in a configuration file.
Given the instruction \(p\), the initial and final workspace states \(s_0, s_H\), and the optional final response \(y\), each verification point returns a binary score:
\begin{equation}
b_i = \mathbb{I}\!\left[c_i(p, s_0, s_H, y)=\mathrm{true}\right],
\qquad b_i \in \{0,1\}.
\end{equation}
The code-based verification score is then computed as the normalized average over all objective checks:
\begin{equation}
s_{\mathrm{code}} = \frac{1}{m}\sum_{i=1}^{m} b_i.
\label{eq:code_score}
\end{equation}
This score measures how well the agent satisfies the deterministic requirements of the task, including both explicitly stated output constraints and implicit conditions derived from task resources.

\paragraph{Rubric-Based Verification.}

Some task requirements cannot be reliably judged by executable code alone. 
For example, a task may require a message to be professional in tone, a report to be concise and well organized, or a summary to faithfully preserve key information without overclaiming. 
To evaluate such qualitative aspects, we generate a set of rubric rules:
\begin{equation}
\mathcal{R} = \{r_1, r_2, \dots, r_n\}.
\end{equation}
Each rule \(r_j\) defines a specific quality dimension and assigns an ordinal score:
\begin{equation}
q_j =
\mathrm{score}_j\!\left[
r_j(p, s_0, s_H, y)
\right],
\qquad
q_j \in \{0, 0.25, 0.5, 0.75, 1.0\},
\end{equation}
where \(r_j\) denotes the \(j\)-th rubric criterion and \(\mathrm{score}_j(\cdot)\) maps the verifier's judgment to one of the predefined discrete scores.
The rubric-based score is computed as a normalized weighted average:
\begin{equation}
s_{\mathrm{rubric}} =
\frac{\sum_{j=1}^{n} w_j q_j}{\sum_{j=1}^{n} w_j},
\label{eq:rubric_score}
\end{equation}
where \(w_j\) denotes the weight of rubric rule \(r_j\). 
By default, all rubric rules use equal weights unless task-specific weighting is required. 
This score captures qualitative completion quality that complements deterministic code checks.

\paragraph{Score Aggregation.}\label{sec:score_agg}

The final task score depends on the available verification signals. 
For tasks with only code-based verification, the final score is directly given by:
\begin{equation}
s_{\mathrm{task}} = s_{\mathrm{code}}.
\end{equation}
For tasks with both code-based and rubric-based verification, we combine the two scores as:
\begin{equation}
s_{\mathrm{task}}
=
\lambda s_{\mathrm{code}}
+
(1-\lambda)s_{\mathrm{rubric}},
\label{eq:task_score}
\end{equation}
where \(\lambda \in [0,1]\) controls the relative importance of objective and subjective verification. 
In our implementation, we set \(\lambda = 0.7\) for hybrid verification, assigning greater weight to code-based checks. 
This choice is justified because Claw task completion is often most reliably evidenced by concrete and reproducible workspace changes, such as generated files, structured outputs, and data transformations. 
The rubric component, weighted by \(0.3\), complements these deterministic checks by assessing qualitative requirements that are difficult to formalize in code. 
This hybrid design prioritizes objectively verifiable task outcomes while still covering higher-level quality requirements in workspace-grounded evaluation.


\subsection{Automated Quality Assessment}
\label{QualityAssessment}
After task synthesis and verification design, we perform automated quality assessment to filter low-quality samples before using them for training or evaluation. This stage evaluates two aspects: the quality of the synthesized task and the reliability of its verification artifacts. Task quality focuses on whether the task is reasonable, self-contained, executable, and clearly specified, while verification quality examines whether the generated code checker and rubric rules can correctly and comprehensively assess task completion. The latter is especially challenging for Claw-style tasks, whose outcomes are often heterogeneous workspace-grounded artifacts, such as modified files, structured data, generated reports, scripts, or user-facing messages, rather than short final answers. Moreover, unlike software-engineering tasks with manually curated reference implementations or test suites from GitHub, Claw tasks require task-specific verifiers to be synthesized automatically. Based on this observation, we design a fine-grained assessment procedure that jointly examines task validity and verification reliability, aiming to identify ambiguous tasks, misaligned checkers, and unreliable evaluation signals.

\subsubsection{Task Quality Assessment}

Task quality assessment aims to identify synthesized tasks that are redundant, infeasible, or poorly calibrated. We mainly assess each task from three aspects: novelty, reasonableness, and difficulty. These dimensions jointly measure whether the task adds new coverage, can be executed in a practical Claw-style workspace, and has an appropriate complexity level.

\paragraph{Novelty.}
We first measure task novelty using cosine similarity between task embeddings. For a newly generated task, we compute its maximum similarity to those in the retained task pool. If the similarity score is below a threshold, the task is considered sufficiently novel and retained; otherwise, it is treated as overly similar and filtered out. This step helps reduce redundant task patterns and ensures that large-scale synthesis improves not only dataset size, but also coverage of distinct task scenarios.

\paragraph{Plausibility.}
We use a powerful LLM (\ie GPT-5.4~\citep{openai2026gpt54}) judge to assess task plausibility as a binary decision. The judge determines whether each task is clear, internally consistent, and realistic as a Claw-style user request. In particular, it checks whether the task relies on unavailable or imaginary environment components, such as non-existent system tools, unsupported services, or unrealistic software integrations. Tasks with plausibility issues are removed to avoid unnatural, unstable, or poorly grounded samples.

\paragraph{Difficulty.}
We also use an LLM (\ie GPT-5.4) judge to estimate task difficulty. This assessment reflects the expected completion complexity, including the number of required steps, operation diversity, and the amount of planning or cross-resource reasoning required. Difficulty scores characterize the task distribution and help maintain a balanced mixture of simple, moderate, and challenging samples.

\subsubsection{Verification Quality Assessment}
We further assess the generated verification artifacts, including code-based and rubric-based checkers. This step is critical because the verifier directly determines how task completion is measured. An incorrect checker invalidates evaluation, an overly lenient checker may inflate scores by rewarding superficial outputs such as merely creating a file, and an overly strict checker may penalize valid solutions by enforcing requirements beyond the task. To address these issues, we apply automated checks for verifier executability and alignment with task requirements. For code-based verification, we examine whether the checker runs correctly, covers core objective requirements, and avoids being too permissive or too restrictive. For rubric-based verification, we check whether they complement the code checker by targeting qualitative aspects difficult to verify deterministically, rather than duplicating code-verifiable requirements.

\paragraph{Code Assessment.}
For code-based verification, we assess both the executability and task alignment of the generated checker. 
1) \textbf{Executable sanity check.} We first ensure that the checker is syntactically valid and can run without errors. To detect overly permissive checkers, we reconstruct the task's initial workspace by placing the required mock input files at their specified paths, but without providing any agent-produced outputs. The checker is then executed on this initial state. If it already assigns a non-trivial score before the task is attempted, the checker is considered unsuitable, since it may reward superficial conditions or fail to distinguish task completion from the initial workspace state. 
2) \textbf{Task-checker alignment review.} We then use a strong LLM judge~(\ie GPT-5.4) to review the checker against the task instruction and input files along two dimensions: requirement coverage and over-strictness. Requirement coverage measures whether the checker verifies the core objective requirements of the task, including both explicit instructions and implicit constraints derived from the input files. Over-strictness measures whether the checker enforces conditions beyond the task specification, such as unnecessary formatting, exact wording, or extra fields. A high-quality checker should cover as many objective task requirements as possible while avoiding constraints not required by the task.

\paragraph{Rubric Assessment.}
For rubric-based verification, we use a strong LLM judge~(\ie GPT-5.4) to assess whether the generated rubric provides complementary evaluation signals to the code checker. Since code-based verification already covers deterministic and objectively checkable requirements, the rubric should focus on qualitative aspects that are difficult to verify through executable checks, such as tone, clarity, organization, faithfulness, and completeness of explanation. The judge reviews the task instruction, input files, code checker, and rubric rules to identify whether any rubric rule merely duplicates code-verifiable requirements. Such overlap can overweight simple objective constraints and reduce the value of the rubric component. A high-quality rubric should therefore complement the checker by covering subjective or high-level quality dimensions that are important for task completion but not reliably captured by code.





\begin{figure}[t]
    \centering
    \begin{subfigure}{0.48\linewidth}
        \centering
        \includegraphics[width=\linewidth]{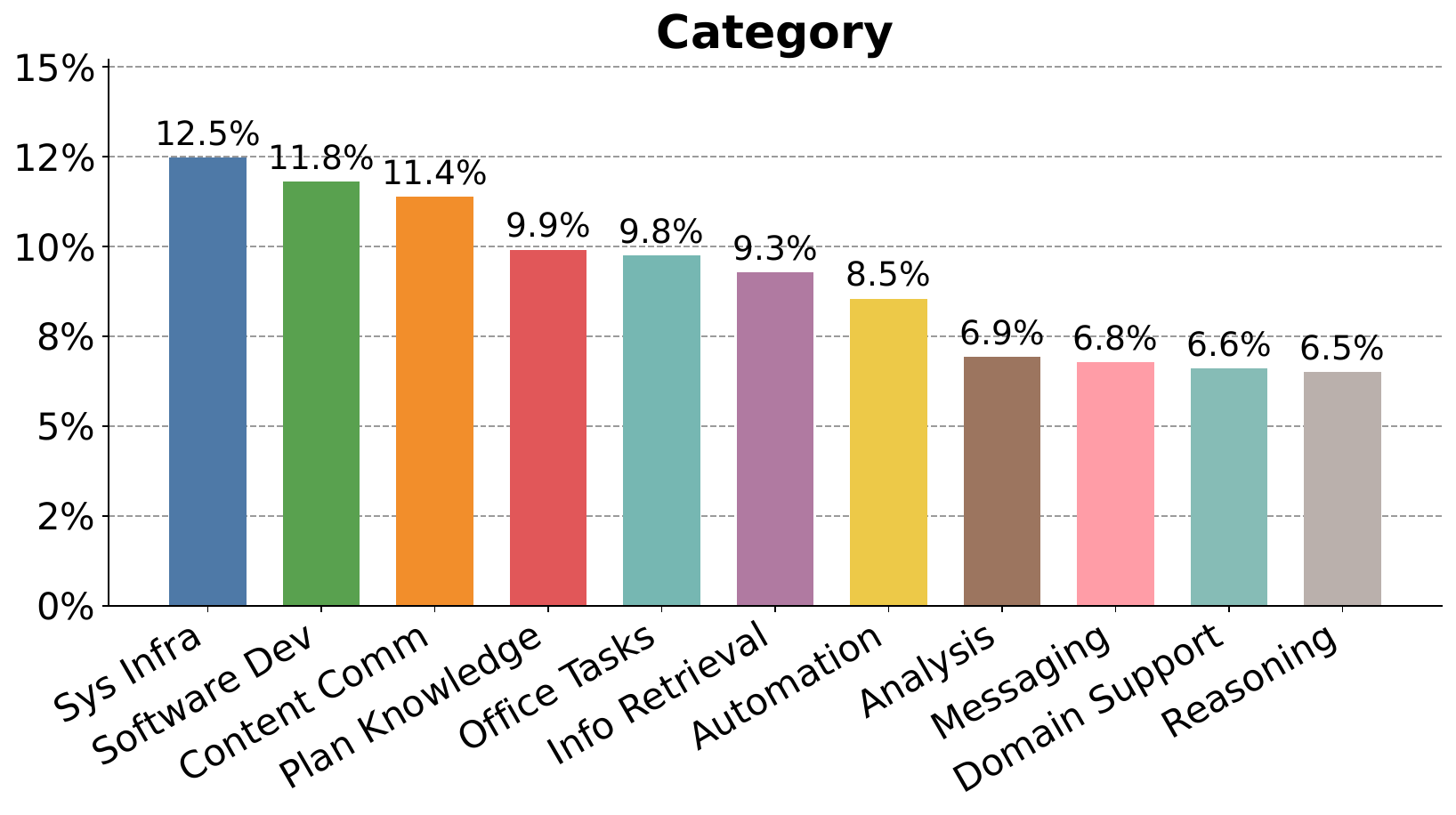}
        \caption{Scenario category distribution.}
        \label{fig:persona_category_distribution}
    \end{subfigure}
    \hfill
    \begin{subfigure}{0.48\linewidth}
        \centering
        \includegraphics[width=\linewidth]{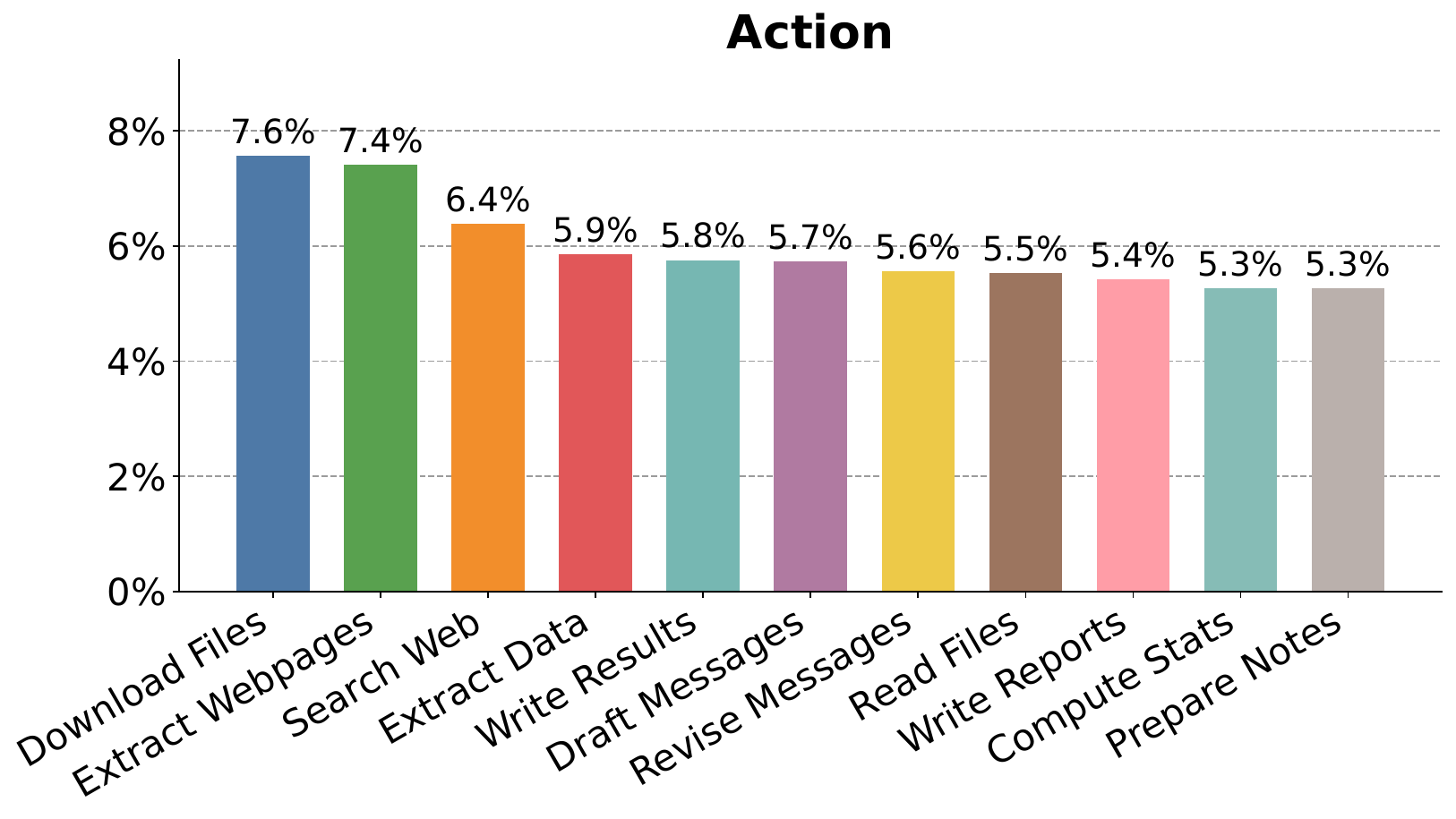}
        \caption{Atomic action distribution.}
        \label{fig:persona_action_distribution}
    \end{subfigure}
    \caption{
    Task distribution of persona-driven synthesis. 
    Figure~(a) shows the distribution of user-facing scenario categories, while Figure~(b) shows the distribution of atomic actions required by the generated tasks. Together, they characterize both the topical diversity and executable-operation diversity of the synthesized tasks.
    }
    \label{fig:persona_task_distribution}
\end{figure}


\begin{table}[t]
    \small
    \centering
    \caption{Category distribution of the annotated skills collected for synthesis.}
    \label{tab:skill_category_distribution}
    \resizebox{0.95\linewidth}{!}{
        \begin{tabular}{cccccccccc}
            \toprule
            \textbf{Category}
            & \textbf{MCP Tools}
            & \textbf{Prompts}
            & \textbf{Workflows}
            & \textbf{Dev Tools}
            & \textbf{Data \& APIs}
            & \textbf{Security}
            & \textbf{Automation}
            & \textbf{Other}
            & \textbf{Total} \\
            \midrule
            \textbf{Count}
            & 411 & 565 & 1972 & 3906 & 4236 & 993 & 1221 & 3533 & 16837 \\
            \textbf{Percentage}
            & 2.44\% & 3.36\% & 11.71\% & 23.20\% & 25.16\% & 5.90\% & 7.25\% & 20.98\% & 100.00\% \\
            \bottomrule
        \end{tabular}
    }
\end{table}

\subsection{Synthesized Task Analysis}
\begin{wraptable}{r}{0.45\textwidth}
\vspace{-1em}
\small
\centering
\caption{Human-sampled quality analysis on 50 randomly sampled training tasks. Each metric is scored on a 1--5 scale.}
\begin{tabular}{lc}
\toprule
Metric & Avg Score \\
\midrule
Task Reasonableness & 4.46 \\
Execution Feasibility & 3.50 \\
Resource Consistency & 4.36 \\
Verification Quality & 3.92 \\
\midrule
Overall & 4.06 \\
\bottomrule
\end{tabular}
\label{tab:human_sampled_task_quality}
\vspace{-1em}
\end{wraptable}
\paragraph{Persona-Driven and Skill-Grounded Task Distribution.}
We analyze the task distributions produced by the two synthesis approaches to examine their coverage and diversity. 
For persona-driven synthesis, Figure~\ref{fig:persona_task_distribution} reports the distributions of scenario categories and atomic actions. 
The generated tasks cover a broad range of user-facing scenarios, with no single category dominating the distribution; even the largest category accounts for only 12.5\% of the tasks. 
The atomic-action distribution also shows diverse executable operations, including information acquisition, workspace inspection, data extraction, computation, message drafting, and report writing. 
These results suggest that persona-driven synthesis produces tasks that are diverse in both scenario coverage and required operations.
For skill-grounded synthesis, we analyze the 16K synthesizable skills retained after filtering. Table~\ref{tab:skill_category_distribution} summarizes the category distribution of the annotated skills used for synthesis. 
The skills cover diverse capability categories, with Data \& APIs and Dev Tools forming the largest groups, followed by Workflows, Automation, Security, Prompts, and MCP Tools. 
This distribution indicates that skill-grounded synthesis is anchored in a broad executable capability space rather than a narrow set of operations. 
Together, the two synthesis approaches are synergistic: persona-driven synthesis expands user-scenario diversity, while skill-grounded synthesis strengthens operational grounding through reusable skills.


\paragraph{Human-Sampled Task Quality Analysis.}

To assess the quality of the synthesized training data, we randomly sample 50 task instances and conduct human evaluation along four dimensions: task reasonableness, execution feasibility, resource consistency, and verification quality. 
These dimensions assess whether the task is clear and realistic, executable under the provided workspace, consistent with its mock resources, and paired with an appropriate verifier. 
Each dimension is scored on a 1--5 scale, and the results are reported in Table~\ref{tab:human_sampled_task_quality}.
The sampled tasks receive positive ratings across both task-side and verifier-side dimensions, with an overall average score of 4.06. 
Task reasonableness and resource consistency score particularly high, indicating that the sampled tasks are generally coherent and well matched with their resources. 
Execution feasibility is relatively lower but still positive, reflecting the non-trivial environment assumptions and multi-step execution complexity of some tasks. 
Overall, these results suggest that the synthesized data maintains a reasonable level of coherence, executability, and verifiability, making it suitable for constructing agent-training trajectories.

\section{ClawGym-Agents: Training Claw Agents with Synthesized Tasks}\label{sec:agents}

This section details the process of leveraging ClawGym-SynData to construct high-quality training trajectories, which are subsequently used to develop the ClawGym-Agents. Given the task specifications and the automatically instantiated workspaces introduced in previous sections, we first collect realistic agent-environment interaction traces through large-scale black-box rollouts on OpenClaw. We then aggregate and clean the raw interaction logs, select high-quality trajectories according to verifier scores, and finally use the resulting dataset to conduct multi-turn SFT. 

\subsection{Black-Box Rollout}

The training paradigm for Claw agents diverges from that of traditional search~\citep{song2025r1_2, li2025webthinkerempoweringlargereasoning} or coding~\citep{sun2026swe, song2026swe} agents. While conventional agentic training~\citep{song2025r1} relies on explicit agentloops that govern agent-environment interactions in a predefined format, frameworks like OpenClaw operate as highly encapsulated systems. Consequently, internal execution details, such as context management and subagent sessions, remain completely opaque.
To collect authentic OpenClaw trajectories at scale, we adopt a black-box rollout strategy that directly leverages the OpenClaw harness. Rather than implementing a separate agent loop or constructing an external harness to approximate OpenClaw's interaction protocol, we execute tasks through OpenClaw itself, thereby preserving its original control flow, tool interface, and agent-environment interaction semantics. Specifically, we deploy multiple OpenClaw Docker environments on a distributed cluster and treat each instance as an executable black-box system. For each synthesized task, we assign the task specification and its associated workspace materials to a designated Docker container, where the initial environment is constructed and the task is solved by invoking the OpenClaw runtime.

During execution, OpenClaw controls the agent to interact with the local environment through tool calls and textual observations. To record these interactions without modifying the agent logic, we introduce a proxy layer that intercepts each request and response exchanged during the rollout. This proxy captures the complete multi-turn interaction stream, including model inputs, model outputs, tool invocations, and environment feedback. As a result, we obtain OpenClaw rollout trajectories that faithfully reflect how the agent solves tasks in realistic local environments. These raw trajectories serve as the foundation for subsequent data aggregation, filtering, and training. 
We leverage MiniMax-M2.5~\citep{minimax_m25} and GLM-5.1~\citep{glm_51} as teacher models to conduct rollouts, constructing a  training corpus by mixing trajectories distilled from both models for subsequent data processing and fine-tuning.

\subsection{Trajectory Aggregation and Selection}\label{sec:traj_agg_and_select}

After collecting black-box rollout logs, we aggregate raw interaction records into complete trajectories. Since the proxy captures requests at the granularity of individual model calls, a single task execution may correspond to multiple partially overlapping request traces. We reconstruct trajectories by grouping requests that share identical message prefixes and concatenating  subsequent interaction turns, thereby recovering coherent multi-turn sequences from fragmented OpenClaw logs. 


In practice, we observe that long-running OpenClaw sessions may explicitly insert auxiliary prompts, such as cron or heartbeat messages, into the conversation. These prompts are used to maintain session activity or initiate secondary interaction branches, but they do not directly contribute to solving the target task. Therefore, during aggregation, we remove such systematic reminders and retain only the interaction segments that are relevant to task execution. In addition, we filter out trajectories involving unsupported tools. For example, some tools, such as canvas-related operations, are not available on the remote rollout infrastructure and are not required by our synthesized tasks. Removing trajectories containing such tool calls prevents spurious tool-call failures from introducing unnecessary noise into the training data.

We then perform trajectory selection based on verifier scores. Unlike conventional program-repair or question-answering tasks, where rewards are often binary, our hybrid verification protocol produces continuous scores in $[0,1]$ by combining code-based and rubric-based verifiers. Consequently, standard best-of-$N$ filtering strategies~\citep{song2026swe} designed for binary success signals may become directly inapplicable. We therefore adopt a simple yet robust reward-threshold-based filtering strategy: after ensuring that a trajectory corresponds to a valid and complete interaction process, we retain it for training only if its final score exceeds a predefined reward threshold. This selection rule prioritizes trajectories that demonstrate sufficiently high task completion quality while avoiding excessive dependence on a binary success criterion, ultimately yielding a final dataset of 24.5K high-fidelity interaction trajectories for SFT. Section~\ref{sec:reward_based_filter_analysis} further analyzes the impact of different reward thresholds on the downstream performance of the trained agents.


After reward-threshold-based filtering, we summarize the interaction and tool-use statistics of the selected trajectories. 
As shown in Table~\ref{tab:trajectory_statistics}, each selected trajectory contains 13.00 interaction rounds and 18.67K tokens on average, indicating that the retained data consists of non-trivial multi-turn executions rather than short single-turn responses. 
The selected trajectories also include 15.82 tool calls and 3.25 distinct tool types on average, suggesting frequent workspace interaction and coordination across multiple operation types. 
These statistics show that the filtered trajectories provide rich Claw-style supervision involving planning, workspace inspection, tool execution, and observation-driven adjustment.

\begin{table}[t]
\small
\centering
\caption{Interaction and tool-use statistics of selected trajectories after filtering. These trajectories are generated using both Minimax-M2.5 and GLM-5.1.}
\begin{tabular}{lcccc}
\toprule
  & Avg. Rounds & Avg. Tokens & Avg. Tool Calls & Avg. Tool Types \\
\midrule
Overall & 13.00 & 18.67K & 15.82 & 3.25 \\
\bottomrule
\end{tabular}
\label{tab:trajectory_statistics}
\end{table}

\subsection{Agentic Training for Claw Agents}

Using the filtered trajectories, we perform multi-turn SFT on the {Qwen3}-series~\citep{yang2025qwen3technicalreport}, including {Qwen3-4B-2507-Instruct}, {Qwen3-8B}, and {Qwen3-30B-A3B-2507-Instruct}. Since Claw tasks often require long-horizon interaction with local files, tools, and environment feedback, the resulting trajectories can substantially exceed the context length used in standard instruction tuning. For {Qwen3-8B}, whose native context window is 32K tokens, we apply YaRN~\citep{peng2023yarnefficientcontextwindow} to extend the maximum context length to 64K tokens, enabling the model to better capture long-range dependencies across multi-turn agent trajectories.

During training, we adopt a multi-turn loss masking strategy tailored to agentic interaction data. Specifically, tokens corresponding to environment feedback produced by Docker-based execution are excluded from the supervised loss. This prevents the model from learning to imitate deterministic execution outputs or tool-returned observations, which are not actions generated by the policy. Instead, the optimization objective focuses on the model-generated parts of the trajectory, including reasoning, decision making, and tool-call generation. This design encourages the model to learn how to act within the environment rather than memorize the environment's responses.

Through this procedure, we obtain \textbf{ClawGym-Agents} trained on our synthesized high-quality task trajectories, namely {ClawGym-4B}, {ClawGym-8B}, and {ClawGym-30B-A3B}. These  trained models possess improved capabilities when deployed to serve as Claw agents.

\subsection{Reinforcement Learning with Sandbox-Parallel Rollouts}
\label{sec:clawgym-rl}

Beyond supervised fine-tuning, we further explore reinforcement learning for Claw-style agents through a lightweight sandbox-parallel pipeline. The pipeline keeps the OpenClaw agent loop as a black box and virtualizes each task into an independent sandbox containing its own root filesystem, workspace, gateway, and verifier~\cite{cheng2026computerenvironmentselicitgeneral}. Our design has three practical advantages: (i) \textit{sandbox-parallel rollout}, which allows many task rollouts to run concurrently without interfering with each other; (ii) \textit{low infrastructure dependency}, enabled by interchangeable Docker-based and Docker-free sandbox backends for different cluster settings; and (iii) \textit{outcome-reward-only training}, where the code verifier provides the reward signal directly, avoiding auxiliary reward models or process-level supervision. 

We conduct RL from two starting points with different capabilities: Qwen3-4B-2507-Instruct, a vanilla instruct model without ClawGym fine-tuning, and ClawGym-30B-A3B, a SFT model initialized from Qwen3-30B-A3B-2507-Instruct, using ClawGym-SynData. For training, we sample 2,000 tasks from ClawGym-SynData and balance the selection to preserve diversity across required tools. We optimize the policy with GRPO~\cite{deepseekmath} using the following hyperparameters: learning rate 1e-6, train batch size 8, rollouts per prompt 8, training steps 100, rollout temperature 0.7, maximum response length 64K tokens. 

\begin{figure}[htb]
	\centering
	\includegraphics[width=0.9\linewidth]{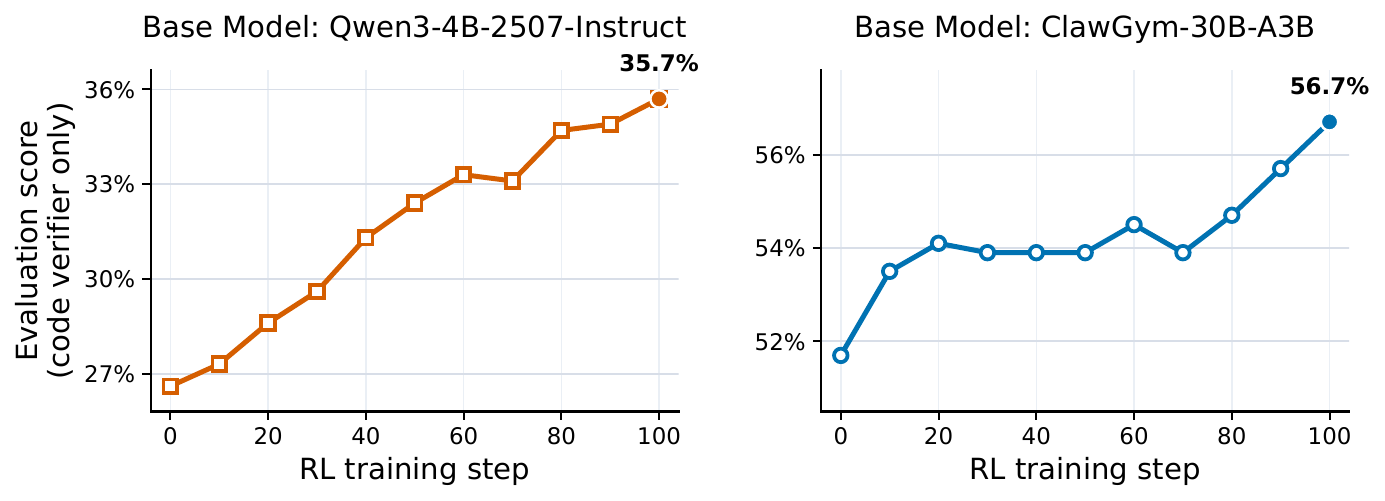}
	\caption{RL training curves on ClawGym-Bench. Scores are computed using only code-based verifiers with weight 1.0, without rubric-based judgment.}
	\label{fig:rl-clawbench-code-verifier}
    \vspace{-10pt}
\end{figure}

Across both starting points, our RL training improves evaluation performance over the corresponding baseline. The gain on the vanilla 4B model shows that RL training can improve Claw-style agents even without prior SFT, while the improvement on ClawGym-30B-A3B indicates that the same recipe remains useful after supervised agent training. These results suggest that lightweight sandbox-parallel rollouts provide a practical RL path for further improving Claw-style agents. The training curves are shown in Figure~\ref{fig:rl-clawbench-code-verifier}.

\section{ClawGym-Bench: A Reliable Benchmark for Claw Agents}
\label{sec:bench}


Reliable evaluation is as important as rigorous training for developing Claw agents. However, it requires more stringent quality control, as it directly measures and compares different models through accurate performance scores. Therefore, we construct \textbf{ClawGym-Bench} from the synthesized task pool through a rigorously designed multi-stage selection process.
This process reduces noisy samples and yields a more reliable benchmark for evaluating Claw agents on workspace-grounded tasks, including file operations, data processing, document editing, code execution, and other realistic computer-use workflows.

\subsection{Benchmark Construction}
We construct ClawGym-Bench from the synthesized task pool after the general task-quality and verifier-quality assessments described above. Since benchmark samples are used for evaluation rather than large-scale training, we apply a stricter construction process that focuses on difficulty calibration, verifier reliability, and coverage of diverse workspace-grounded scenarios.

\paragraph{Difficulty-Aware Filtering.}
We denote by \(\mathcal{D}_{\mathrm{cand}}\) the set of synthesized tasks that remain after task-quality and verifier-quality filtering in Section~\ref{QualityAssessment}. To further select suitable benchmark samples, we perform rollout-based difficulty filtering on this candidate pool. For each task \(\tau \in \mathcal{D}_{\mathrm{cand}}\), we run \(n=4\) rollouts with both a strong LLM agent and a smaller LLM agent, and compute their average completion scores as \(\bar{s}_{\mathrm{strong}}(\tau)\) and \(\bar{s}_{\mathrm{small}}(\tau)\), respectively. These scores provide an empirical estimate of whether the task is solvable, whether it is too easy, and whether it can distinguish agents with different capabilities. 
Specifically, a task is retained only if it satisfies the following criteria:
\begin{equation}
\left\{
\begin{aligned}
\bar{s}_{\mathrm{strong}}(\tau) &\geq 0.2, \\
\bar{s}_{\mathrm{small}}(\tau) &\leq 0.6, \\
\bar{s}_{\mathrm{strong}}(\tau) &> \bar{s}_{\mathrm{small}}(\tau).
\end{aligned}
\right.
\label{eq:difficulty_filter}
\end{equation}

The first condition removes tasks that are too difficult or unstable even for the strong agent, while the second filters out tasks that are already easy for the smaller agent. The third condition ensures that the strong agent performs better than the smaller one, indicating that the task reflects a meaningful capability gap. This rollout-based filtering helps ClawGym-Bench avoid both trivial and unrealistically difficult samples, while retaining tasks that are suitable for discriminative evaluation.

\paragraph{LLM-Assisted Human Review.}
After automated filtering and difficulty-aware selection, the remaining candidates undergo a final human-LLM collaborative review before being included in the benchmark. 
A Claw benchmark instance consists of multiple interdependent components, including the task instruction, input resources, executable checkers, and optional rubric rules. 
Reviewing all these components purely manually is inefficient and may still overlook subtle issues, especially when checker logic depends on constraints hidden in the input files or when task-verifier alignment requires code-level reasoning.
To make this stage more effective, we use a frontier LLM (\ie GPT-5.4) to perform a detailed diagnostic review. 
The model examines the task instruction, input files, code checker, and rubric rules, then reports potential problems and suggests concrete revisions. 
Its review focuses on whether the task is clear and feasible, whether the input resources support the intended task, whether the checker is executable and appropriately strict, and whether the rubric complements rather than duplicates code-based verification. 
Human reviewers then act as final decision makers: they examine the LLM feedback, determine whether the identified issues are valid, decide whether to accept, revise, or reject each candidate, and add missing details when necessary. 
This review process combines the diagnostic strength of frontier LLMs with human oversight, improving review efficiency while keeping final decisions under human control.

\paragraph{Benchmark Composition.}

The final benchmark contains 200 tasks retained after difficulty-aware filtering and Human-LLM review. Each benchmark instance includes a user instruction, task-specific mock resources, and the corresponding verifier. Among these tasks, 156 are evaluated purely by code-based checkers, while 44 use hybrid verification that combines code-based checks with rubric-based judgment for qualitative requirements. We compute the final score of each task using the verification aggregation defined in Equation~\ref{eq:task_score}. The benchmark covers 6 task categories, with the category definitions and proportions summarized in Table~\ref{tab:bench_category_distribution}. This composition allows ClawGym-Bench to evaluate agents across a diverse set of workspace-grounded scenarios, including file manipulation, data analysis, document editing, script execution, and communication-oriented workflows.

\begin{table*}[t]
\small
\centering
\caption{Category distribution of ClawGym-Bench.}
\begin{tabularx}{0.95\textwidth}{@{}lXr@{}}
\toprule
Category & Description & \#Tasks \\
\midrule
Productivity and Collaboration 
& Routine office work, messaging, scheduling, coordination, and \mbox{collaborative} workflows. 
& 44 \\
\midrule
Systems and Automation 
& System inspection, workflow automation, operational \mbox{execution}, and environment-level handling. 
& 42 \\
\midrule
Analysis and Reasoning 
& Data analysis, evaluation, decision support, problem solving, and \mbox{technical} reasoning. 
& 35 \\
\midrule
Content and Domain Support 
& Professional-domain assistance, external-facing writing, \mbox{reporting}, and communication tasks. 
& 28 \\
\midrule
Planning and Knowledge 
& Information collection, knowledge organization, retrieval, \mbox{synthesis}, and planning-oriented work. 
& 26 \\
\midrule
Software Development 
& Code implementation, debugging, testing, refactoring, and \mbox{software} engineering workflows. 
& 25 \\
\midrule
\textbf{All}
& 
& \textbf{200} \\
\bottomrule
\end{tabularx}
\label{tab:bench_category_distribution}
\end{table*}

\subsection{Evaluation Protocol}

ClawGym-Bench evaluates agents in a harness-based workspace setting: each task is initialized with its mock resources, the agent interacts with the workspace, and the final state is scored by the corresponding verifier using Equation~\ref{eq:task_score}. Since such execution-based evaluation is substantially more expensive than text-only scoring, the benchmark is designed to emphasize two properties that directly affect evaluation reliability: \emph{evaluation stability} and \emph{verifiable solvability}.




\paragraph{Evaluation Stability.}
A practical benchmark for Claw agents should yield stable results under repeated harness-based evaluation. This property is particularly important because each Claw evaluation involves multi-turn interaction in the harness, long execution contexts, and frequent tool calls, making repeated evaluation substantially more costly than text-only benchmarking. To examine stability, we evaluate models of different sizes on a fixed subset of 50 benchmark tasks, sampled in a category-balanced manner to cover diverse task types. Each model is evaluated 5 times under the same harness setting, with identical workspace initialization, mock resources, and verification artifacts. As shown in Table~\ref{tab:evaluation_sta}, the standard deviations across repeated runs are small~($\le$ 1\%), indicating that ClawGym-Bench supports reliable model comparison without requiring many repeated runs solely to average out randomness.

\begin{table}[htbp]
\small
\centering
\caption{Stability analysis under repeated harness-based evaluation.}
\begin{tabular}{lccc}
\toprule
Model & Runs & Mean Score(\%) & Std.(\%) \\
\midrule
Qwen3-8B & 5 & 36.4 & 0.3 \\
Qwen3-30B-A3B & 5 & 42.6 & 1.0 \\
\bottomrule
\end{tabular}
\label{tab:evaluation_sta}
\end{table}

\paragraph{Verifiable Solvability.}
Each benchmark task is checked to ensure that it admits at least one feasible path to the full score. This property is important because a task should not be impossible to fully solve due to flaws in the task instruction, missing or inconsistent input resources, or overly strict verification code. During benchmark construction, we verify this property either through successful strong-agent rollouts or through human-constructed reference completions. In both cases, the final workspace state is evaluated by the corresponding verifier to confirm that the task can receive full credit. This check helps ensure that failures on ClawGym-Bench reflect limitations of the evaluated agent rather than artifacts of infeasible task design, invalid input files, or unreachable verification.

\section{Experiment}
\label{exp}

In this section, we evaluate the proposed ClawGym framework by extensively examining the data quality, training effectiveness, and benchmark reliability.

\subsection{Experimental Settings}

\paratitle{Evaluation Datasets and Metrics.}
We evaluate all models on ClawGym-Bench and PinchBench~\citep{pinchbench2026} using a hybrid evaluation protocol that combines code-based verification with rubric-based judgment. For ClawGym-Bench tasks whose verifier contains only code checks, we directly use the corresponding code-based verifier for automated assessment, with a weight of 1.0; this score is taken as the final task score. For tasks whose verifier contains both code checks and rubrics, we assign a weight of 0.7 to the code-based component and 0.3 to the rubric-based component, as described in Section~\ref{sec:score_agg}. For PinchBench, we use the task set released on April 10, 2026, excluding multimodal tasks and retaining 30 tasks in total. Each PinchBench task is evaluated using the verifier provided in the original benchmark specification. The rubric-based judgments are conducted using GPT-5.4 as the judge model, and the corresponding prompts are provided in Appendix~\ref{app:prompt}.

\paratitle{Evaluated Models.}
We evaluate a broad set of models spanning both proprietary and open-weight systems. The proprietary frontier models include Claude-4.7-Opus~\citep{anthropic_claude_opus_4_7}, Claude-4.6-Opus~\citep{anthropic_claude_opus_4_6}, Claude-4.6-Sonnet~\citep{anthropic_claude_sonnet_4_6}, Gemini-3-Flash~\citep{google2026gemini3flash}, and GPT-5.4~\citep{openai2026gpt54}. The open-weight frontier models include DeepSeek-V3.2~\citep{liu2025deepseek}, GLM-5.1~\citep{glm_51}, MiniMax-M2.7~\citep{minimax2026m27}, Kimi-K2.6~\citep{kimi_k26}, and Qwen3.5-Plus~\citep{qwen_35_plus}. We further evaluate models from the Qwen3-series~\citep{yang2025qwen3technicalreport}, as well as ClawGym-Agents obtained by using Qwen3-series models as backbone models and fine-tuning them on ClawGym-SynData. For all evaluated models, trajectories are collected through black-box rollouts in the original OpenClaw environments. For the Qwen-series models, we adopt generation configurations tailored to their respective model specifications, while all other models are evaluated through direct API calls. The context length is uniformly set to 64K tokens.

\subsection{Main Results}
We organize the evaluation of ClawGym around three main questions. 
First, we examine whether training on ClawGym-SynData improves compact open-weight models and narrows their gap with stronger agents. 
Second, we assess whether ClawGym-Bench provides a sufficiently discriminative evaluation across model scales and task categories. 
Third, we test whether the resulting ClawGym-Agents generalize beyond the synthesized task distribution by evaluating them on the external PinchBench benchmark. 
Table~\ref{tab:main_results_by_category} summarizes the overall results.

\begin{table*}[ht]
\small
\centering

\caption{
Performance comparison of different models on ClawGym-Bench and PinchBench. The best and second-best Pass@1 results within each model group are highlighted in \textbf{bold} and \underline{underlined}, respectively. Results marked with $^\ast$ are taken directly from the official websites.
}
\footnotesize
\setlength{\tabcolsep}{3.2pt}
\renewcommand{\arraystretch}{1.12}
\newcommand{\NA}{--}

\resizebox{\textwidth}{!}{
\begin{tabular}{@{}l@{\hspace{2pt}}cccccccc@{}} 

\toprule
\multirow{2}{*}[-0.5ex]{\textbf{Model}} 
& \multirow{2}{*}[-0.5ex]{\makecell{\textbf{Pinch-}\\ \textbf{Bench}}}
& \multicolumn{7}{c}{\textbf{ClawGym-Bench}} \\
\cmidrule(lr){3-9}
& 
& \makecell{\textbf{Product.}\\\textbf{\& Collab.}} 
& \makecell{\textbf{Systems}\\\textbf{\& Auto.}} 
& \makecell{\textbf{Analysis}\\\textbf{\& Reason.}} 
& \makecell{\textbf{Content}\\\textbf{\& Domain}} 
& \makecell{\textbf{Planning}\\\textbf{\& Knowl.}} 
& \makecell{\textbf{Software}\\\textbf{Dev.}} 
& \textbf{Avg.} \\
\midrule

\rowcolor{gray!12}
\multicolumn{9}{c}{\textbf{Proprietary Frontier Models}} \\
\addlinespace[2pt]
Claude-4.7-Opus
& \underline{79.40}\hspace*{1.5mm} & \underline{73.02} & \textbf{74.28} & \textbf{80.14} & \textbf{80.98} & \textbf{81.04} & 82.04 & \textbf{77.81} \\
Claude-4.6-Sonnet
& 79.30$^\ast$ & 70.44 & 68.04 & \underline{76.27} & 74.59 & 74.21 & 73.40 & 72.40 \\
Claude-4.6-Opus
& 75.30$^\ast$ & 68.02 & 63.49 & 73.23 & 74.40 & 73.38 & \underline{82.90} & 71.43 \\
Gemini-3-Flash
& \textbf{88.70}$^\ast$ & 68.70 & \underline{68.06} & 68.89 & \underline{78.87} & \underline{80.68} & \textbf{84.04} & \underline{73.50} \\
GPT-5.4
& 68.30$^\ast$ & \textbf{76.00} & 66.56 & 73.83 & 76.19 & 68.51 & 82.40 & 73.49 \\
\addlinespace[2pt]

\rowcolor{gray!12}
\multicolumn{9}{c}{\textbf{Open-Weight Frontier Models}} \\
\addlinespace[1pt]
DeepSeek-V3.2
& 60.80$^\ast$ & 64.45 & \underline{64.43} & 71.97 & 72.30 & 66.66 & 70.34 & 67.90 \\
GLM-5.1
& 76.40\hspace*{1.5mm} & \textbf{70.95} & 64.11 & 71.71 & \textbf{75.99} & \textbf{74.09} & \underline{73.86} & \textbf{71.12} \\
MiniMax-M2.7
& \textbf{80.90}$^\ast$ & 63.98 & 60.11 & 63.96 & 63.42 & 68.72 & 64.10 & 63.72 \\
Kimi-K2.6
& 66.50\hspace*{1.5mm} & 67.16 & 55.99 & \underline{72.56} & 72.84 & 71.56 & \textbf{82.47} & 69.05 \\
Qwen3.5-Plus
& \underline{78.70}\hspace*{1.5mm} & \underline{67.34} & \textbf{66.56} & \textbf{74.27} & \underline{73.07} & \underline{71.79} & 72.10 & \underline{70.35} \\
\addlinespace[2pt]

\rowcolor{gray!12}
\multicolumn{9}{c}{\textbf{Compact Open-Weight Models}} \\
\addlinespace[1pt]
Qwen3-8B
& 54.50\hspace*{1.5mm} & 37.46 & 29.06 & 30.40 & 41.12 & 44.47 & 30.69 & 35.02 \\
Qwen3-32B
& 49.40\hspace*{1.5mm} & 40.68 & 36.21 & 37.84 & 47.16 & 49.29 & 33.11 & 40.32 \\
Qwen3-30A3B
& 55.60\hspace*{1.5mm} & 42.47 & 42.09 & 45.04 & 51.95 & 45.98 & 46.24 & 45.11 \\
Qwen3-235A23B
& 60.60\hspace*{1.5mm} & \textbf{53.66} & \textbf{52.27} & 47.18 & \underline{61.39} & \textbf{67.33} & 49.23 & \underline{54.48} \\
\addlinespace[2pt]

\midrule
\textbf{ClawGym-4B}
& \underline{76.40}\hspace*{1.5mm} & 45.21 & 46.60 & \underline{47.51} & 48.99 & 52.25 & 48.27 & 47.73 \\
\textbf{ClawGym-8B}
& 75.70\hspace*{1.5mm} & 49.47 & 46.83 & 46.35 & 55.37 & 52.29 & \underline{54.78} & 50.24 \\
\textbf{ClawGym-30A3B}
& \textbf{86.00}\hspace*{1.5mm} & \underline{52.98} & \underline{50.97} & \textbf{64.64} & \textbf{61.46} & \underline{57.90} & \textbf{56.13} & \textbf{56.82} \\
\bottomrule
\end{tabular}
}

\label{tab:main_results_by_category}
\end{table*}

\noindent $\bullet$ \emph{Effectiveness of Synthesized Data.} Training on our synthesized data yields substantial performance gains for compact open-weight backbones. As shown in Table~\ref{tab:main_results_by_category}, ClawGym-4B, 8B, and 30A3B achieve average scores of 47.73, 50.24, and 56.82 on ClawGym-Bench, respectively, consistently outperforming their Qwen baselines. Notably, ClawGym-30A3B surpasses the much larger Qwen3-235B-A23B, suggesting that high-quality, agent-specific interaction data can partially compensate for the limitations of model scale. Furthermore, on PinchBench, ClawGym-30A3B achieves a score of 86.00, which is highly competitive even when compared to the scores reported for several proprietary frontier models. These results indicate that our synthesis pipeline effectively equips smaller models with specialized agentic capabilities, narrowing the performance gap between compact and large-scale backbones in these scenarios.


\noindent $\bullet$ \emph{Discriminative Capacity of ClawGym-Bench.}
ClawGym-Bench provides discriminative signals across both task categories and model capability levels. 
Across categories, no single agent dominates all dimensions; while Claude-4.7-Opus achieves the highest overall average, GPT-5.4 excels in \textit{Product. \& Collab.} and Gemini-3-Flash leads in \textit{Software Dev.}
This category-wise variation indicates that different benchmark categories probe distinct agent capabilities and can expose model-specific strengths. 
Regarding model-level performance, the average scores establish a clear hierarchy ranging from 35.02 for Qwen3-8B to 77.81 for Claude-4.7-Opus, which demonstrates a high degree of granularity in distinguishing between models. 
Collectively, these results indicate that ClawGym-Bench is both challenging and discriminative across model scales and task categories.



\noindent $\bullet$ \emph{Generalization beyond Synthesized Scenarios.} The robust performance of ClawGym-Agents on PinchBench demonstrates the strong generalizability of our ClawGym framework. Since these models were supervised exclusively by ClawGym-SynData, their success on an external benchmark indicates that the data effectively facilitates the acquisition of transferable agentic principles. The fact that these gains are not restricted to the synthesized task distribution suggests that the models are internalizing core functional logic rather than merely overfitting to specific patterns within ClawGym-SynData. These results validate our synthesized data as a scalable and high-fidelity supervision source, capable of equipping agents with the robustness required for diverse, real-world computer-use scenarios.

\subsection{Experimental Analysis}
In this section, we conduct a series of empirical analyses to investigate the key factors contributing to the performance of ClawGym-Agents SFT. Specifically, we evaluate the synergy between our dual synthesis strategies, analyze the training dynamics and convergence patterns, and examine the impact of reward-based trajectory filtering protocols.

\paratitle{Synergy of Synthesis Strategies.}
To assess the necessity of persona-driven top-down synthesis and skill-grounded bottom-up synthesis, we compare models trained on trajectories collected from different task sources. \textit{Only Persona-driven} and \textit{Only Skill-grounded} use trajectories from the corresponding strategy alone, while \textit{Mixed Synthesis} combines both sources.
As shown in Table~\ref{tab:synthesis_strategy}, models trained with mixed synthesis outperform those trained with either strategy alone, highlighting the synergy between top-down intent diversity and bottom-up operational grounding. Specifically, the persona-driven strategy ensures a broad coverage of realistic user scenarios, while the skill-grounded strategy anchors tasks in the agent's actual functional boundaries. By combining these, the model learns to bridge the gap between abstract user needs and concrete tool execution, leading to better generalization across diverse tasks.

\begin{table}[ht]
\centering
\footnotesize
\setlength{\tabcolsep}{6pt}
\renewcommand{\arraystretch}{1.12}
\caption{
Synergy between Persona-Driven and Skill-Grounded Synthesis Strategies.
}
\begin{tabular}{llcc}
\toprule
\textbf{Base Model}
& \textbf{Training Data Source}
& \textbf{ClawGym-Bench}
& \textbf{PinchBench} \\
\midrule
\multirow{3}{*}{Qwen3-8B}
& \textit{Only Persona-driven} & 49.44 & 73.51 \\
& \textit{Only Skill-grounded} & 49.06 & 68.23 \\
& \textit{Mixed Synthesis} & 50.24 & 75.68 \\
\midrule
\multirow{3}{*}{Qwen3-30A3B}
& \textit{Only Persona-driven} & \underline{53.65} & \underline{84.92} \\
& \textit{Only Skill-grounded} & 52.27 & 80.05 \\
& \textit{Mixed Synthesis} & \textbf{56.82} & \textbf{86.00} \\
\bottomrule
\end{tabular}
\label{tab:synthesis_strategy}
\end{table}

\begin{figure}[ht]
\centering
    \centering
    \includegraphics[width=0.8\linewidth]{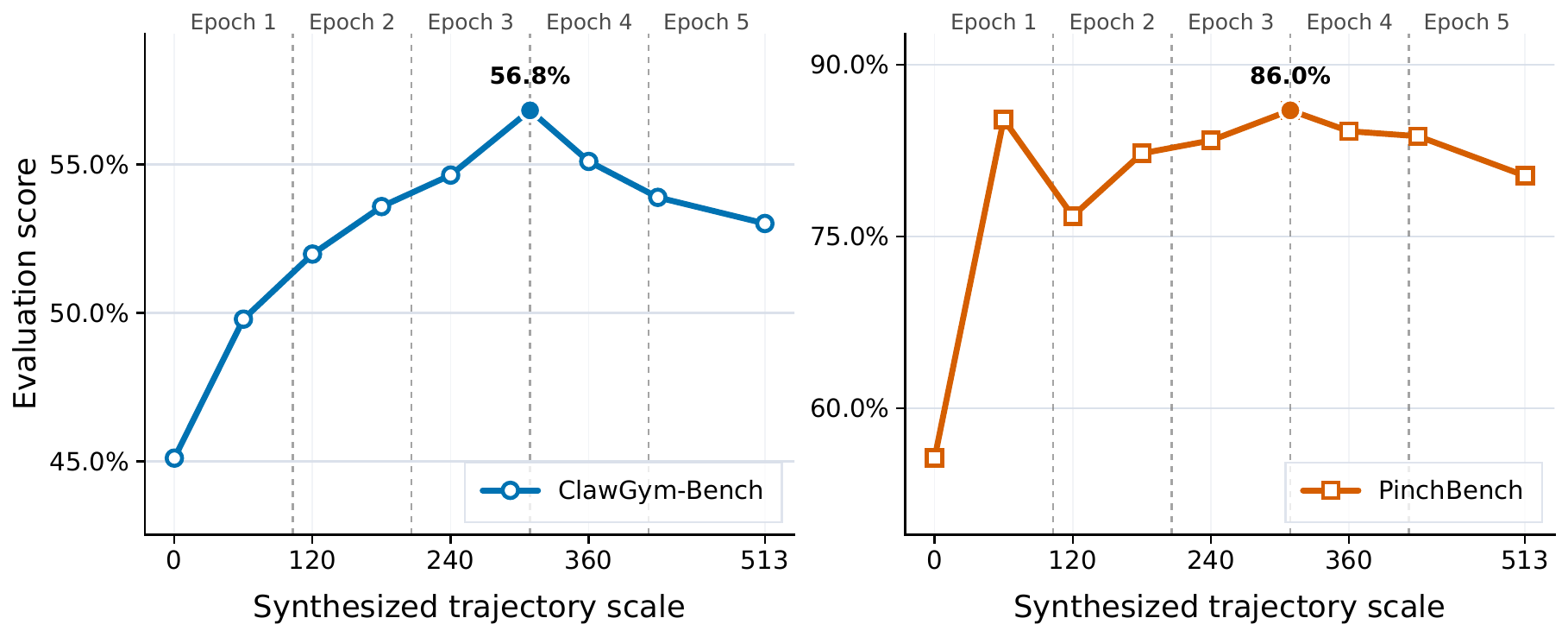}
\caption{Effect of training trajectory scale on SFT Model using ClawGym-SynData.}
    \label{tab:dynamics}
\end{figure}

\paratitle{Training Dynamics and Convergence.}
To investigate how the volume of training exposure affects model performance, we analyze the training dynamics across 5 epochs (103 steps per epoch), evaluating checkpoints every 60 steps. As shown in Figure~\ref{tab:dynamics}, both {ClawGym-Bench} and {PinchBench} exhibit significant gains over the base model (step 0), validating the high quality of our synthesized trajectories. Notably, performance peaks at the conclusion of the third epoch (step 309), after which we observe a slight but steady decline. This trend suggests that identifying the optimal training scale is crucial; while initial exposure allows the model to effectively internalize agent-specific capabilities, excessive iterations lead to marginal over-fitting on the synthesized data distribution.

\begin{figure}[ht]
\centering
    \centering
    \includegraphics[width=0.8\linewidth]{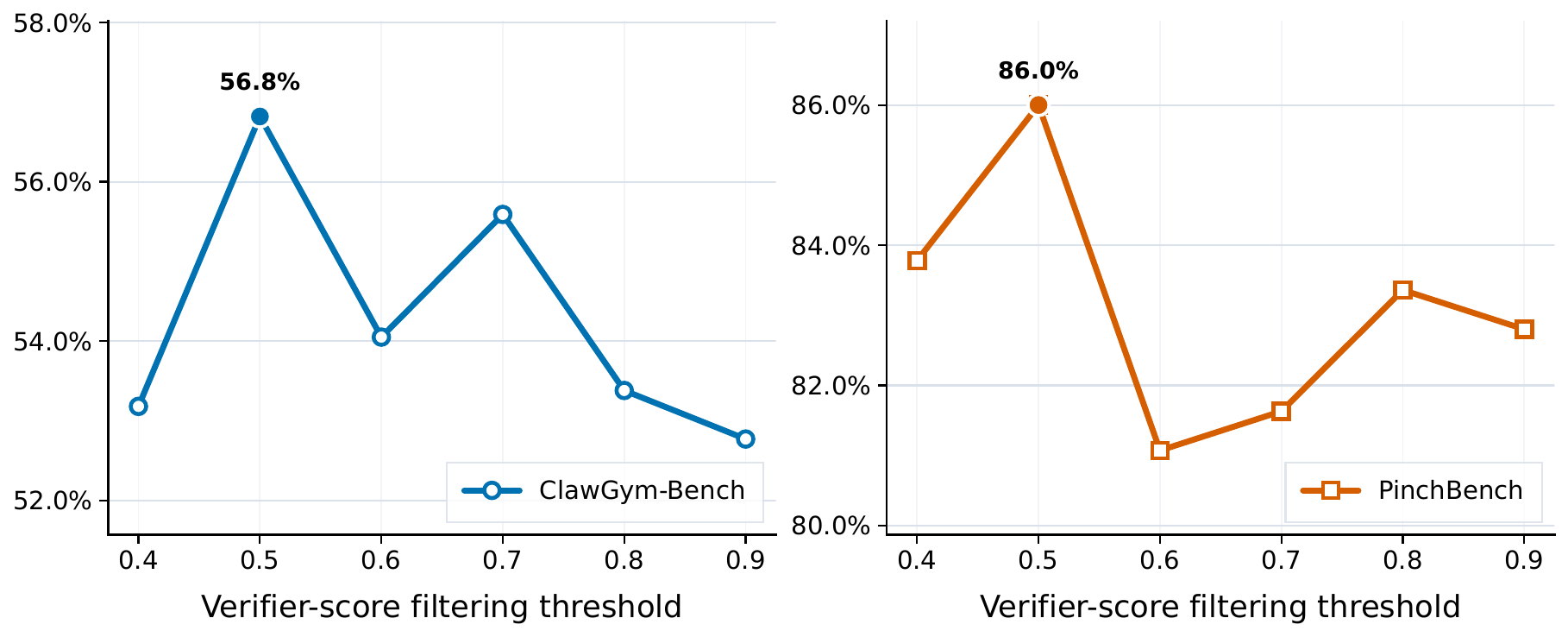}
\caption{Effect of trajectory filtering reward threshold on SFT Model ClawGym-SynData.}
    \label{fig:1_datascaling_rewardchange_rewardchange}
\end{figure}

\paratitle{Impact of Reward Thresholding}\label{sec:reward_based_filter_analysis}
As discussed in Section~\ref{sec:traj_agg_and_select}, we adopt a simple reward-based thresholding strategy to select trajectories for agentic training. Since our hybrid verifier yields continuous scores in $[0,1]$, the choice of threshold directly governs the trade-off between task completion fidelity and data coverage. To identify the optimal threshold, we evaluate performance across a range from 0.4 to 0.9. As shown in 
Figure~\ref{fig:1_datascaling_rewardchange_rewardchange}, a threshold of 0.5 yields the best results on both benchmarks, marking a critical equilibrium between execution completeness and data diversity. While lower thresholds introduce trajectories with insufficient task fulfillment that may degrade the supervision signal, excessively high thresholds prune valuable behavioral variety. Specifically, overly stringent filtering may discard trajectories that, despite imperfect completion scores, demonstrate essential recovery patterns or partial strategies for complex, heterogeneous tasks. These results indicate that calibrated filtering is essential for ensuring high completion fidelity while preserving the diversity necessary for robust agentic generalization. However, the optimal threshold remains an empirical design choice under the current pipeline. In future work, we plan to investigate more principled selection methods for continuous verifier scores.




\section{Behavioral Analysis of Claw Agents}

\label{sec:capability_landscape}

Claw-style tasks require agents to use local files and tools, process execution feedback, and update persistent workspaces. They involve interdependent abilities such as tool-use appropriateness, long-horizon execution robustness, and fine-grained instruction following. We compare GPT-5.4 and Qwen3-30B-A3B~(30A3) on ClawGym-Bench cases to identify effective execution patterns and common failures, offering insights for future data synthesis, training, and evaluation.

\begin{figure}[!t]
\centering
\begin{AIbox}{Case 1: Tool-Use Appropriateness}

\noindent
{\bf Task requirement.}
{\scriptsize
Generate two CI artifact audit reports from all JSON exports under
\texttt{input/ci-artifacts/*.json}. The agent must group artifacts by repository
and artifact name, compute size statistics, count missing sizes and critical
instances, normalize repository identifiers, and write
\texttt{output/audit/report.json} and
\texttt{output/audit/report\_filtered.json}. The filtered report must use the regex
\texttt{(coverage|test-results)}.
}

\vspace{0.6em}

\begin{minipage}[t]{0.49\linewidth}
{\bf Strong Model: Reliable Tool Workflow}

\vspace{0.25em}
{\scriptsize
The higher-scoring trajectory uses tools as a staged computation pipeline. It first
expands the file pattern, then inspects the JSON structure, and finally runs a Python
aggregation script to compute both reports from the underlying records. The printed
summaries expose key quantities required by the verifier.
}

\vspace{0.35em}
{\scriptsize\ttfamily
exec: list input/ci-artifacts/*.json\par
output: gadgets.json, gizmos.json,\par
\quad widgets.json\par
\vspace{0.2em}
\goodhl{exec: inspect JSON shape with python3}\par
output: keys = repository,\par
\quad total\_count, artifacts\par
output: LISTKEY artifacts len 3 / 5\par
\vspace{0.2em}
\goodhl{exec: run Python aggregation script}\par
write output/audit/report.json\par
\goodhl{write output/audit/report\_filtered.json}\par
\vspace{0.2em}
FULL summary:\par
\quad files\_scanned = 3\par
\quad records\_scanned = 11\par
\quad records\_missing\_size = 1\par
\goodhl{\quad groups = 7}\par
\quad critical\_instances = 2\par
\vspace{0.2em}
FILTERED summary:\par
\quad artifact\_match = (coverage|test-results)\par
\goodhl{\quad groups = 3}\par
\quad critical\_instances = 0\par
\textbf{reward = 1.000}\par
}
\end{minipage}
\hfill
\begin{minipage}[t]{0.49\linewidth}
{\bf Weak Model: Brittle Tool Use}

\vspace{0.25em}
{\scriptsize
The weaker trajectory is not a no-op. After an invalid wildcard read fails, it
recovers by using \texttt{find} and reads the three JSON files. However, the later
workflow does not establish a comparably reliable aggregation-and-verification
process, leading to only partial satisfaction of the exact report checks.
}

\vspace{0.35em}
{\scriptsize\ttfamily
read input/ci-artifacts/*.json\par
\badhl{\textbf{Tool error:} ENOENT,}\par
\badhl{\quad wildcard not expanded}\par
\vspace{0.2em}
exec: find input/ci-artifacts\par
\quad -name '*.json' | sort\par
output: gadgets.json, gizmos.json,\par
\quad widgets.json\par
\vspace{0.2em}
read input/ci-artifacts/gadgets.json\par
read input/ci-artifacts/gizmos.json\par
read input/ci-artifacts/widgets.json\par
...\par
write output/audit/report.json\par
write output/audit/report\_filtered.json\par
\vspace{0.2em}
\badhl{\textbf{Unresolved issues:}}\par
\badhl{\quad exact summary fields}\par
\badhl{\quad group ordering / counts}\par
\badhl{\quad filtered report semantics}\par
\badhl{\quad critical-instance checks}\par
\textbf{reward = 0.308}\par
}
\end{minipage}



\vspace{0.3em}
\end{AIbox}

\caption{The stronger trajectory builds a computation-and-verification pipeline, while the weaker one remains brittle in report aggregation despite recovering from an initial tool error.}
\label{fig:case_tool_use_ci_audit}
\end{figure}

\begin{figure}[!t]
\centering
\begin{AIbox}{Case 2: Long-Horizon Execution Robustness}

\noindent
{\bf Task requirement.}
{\scriptsize
Build a local automation for support-ticket batches under \texttt{input/inbox/*.jsonl}.
For each batch, the agent must generate ticket rewrites, batch metrics, team
notifications, and an idempotent state file at
\texttt{out/state/processed\_batches.json}. The automation must process the two
provided examples and remain safe to rerun.
Here, \texttt{batch-04-15} and \texttt{batch-04-16} abbreviate the two provided
support-ticket input files.
}

\vspace{0.6em}

\begin{minipage}[t]{0.49\linewidth}
{\bf Strong Model: Recovery and Closure}

\vspace{0.25em}
{\scriptsize
The stronger trajectory encounters errors, but treats them as recoverable execution
feedback. After missing optional memory files and later hitting an \texttt{exec}
preflight refusal, it switches to a safer workflow, resets only the state file,
reruns the processor, and verifies that the second run does not duplicate work.
}

\vspace{0.35em}
{\scriptsize\ttfamily
read MEMORY.md\par
\goodhl{\textbf{Recoverable error:} ENOENT}\par
\vspace{0.2em}
list input/inbox\par
read batch-04-15.jsonl\par
read batch-04-16.jsonl\par
\vspace{0.2em}
\goodhl{write process\_support\_batches.py}\par
exec: python3 process\_support\_batches.py\par
check generated rewrites/metrics/notifications\par
\vspace{0.2em}
edit: preserve API, UTC, README,\par
\quad CONTRIBUTING, CSV, CLI\par
\vspace{0.2em}
exec: complex cleanup command\par
\goodhl{\textbf{Recoverable error:} preflight refused}\par
\vspace{0.2em}
\goodhl{fallback: write state = []}\par
\goodhl{exec: python3 process\_support\_batches.py}\par
\goodhl{exec: python3 process\_support\_batches.py}\par
read processed\_batches.json\par
\goodhl{output: two batches, no duplicates}\par
\textbf{reward = 1.000}\par
}
\end{minipage}
\hfill
\begin{minipage}[t]{0.49\linewidth}
{\bf Weak Model: Failure Cascades}

\vspace{0.25em}
{\scriptsize
The weaker trajectory does not convert tool failures into a stable recovery plan.
After repeated missing-file errors and an approval-related dead-end, the workflow
fails to converge to a complete runnable automation. As a result, the required
per-batch artifacts and idempotent state tracking are largely missing.
}

\vspace{0.35em}
{\scriptsize\ttfamily
attempt to inspect workspace files\par
\badhl{\textbf{Unresolved error:} repeated missing files}\par
...\par
attempt partial setup actions\par
...\par
\badhl{\textbf{Dead-end:} approval request blocks progress}\par
\badhl{/approve ... allow-once}\par
...\par
\badhl{\textbf{Missing:} complete rewrites}\par
\badhl{\textbf{Missing:} metrics summaries}\par
\badhl{\textbf{Missing:} team notifications}\par
\badhl{\textbf{Missing:} valid idempotent state}\par
\textbf{reward = 0.067}\par
}
\end{minipage}



\vspace{0.3em}
\end{AIbox}

\caption{A representative case of error recovery in long-horizon execution. The stronger trajectory recovers from tool failures and completes an idempotent workflow, while the weaker trajectory accumulates unresolved errors and fails to reach stable completion.}
\label{fig:case_error_recovery_long_horizon}
\end{figure}

\paragraph{Tool-Use Appropriateness.}
In Claw-style environments, task success depends not only on invoking tools, but also on using them coherently within an execution workflow. Figure~\ref{fig:case_tool_use_ci_audit} illustrates this distinction in a CI artifact audit task. The stronger trajectory expands the file pattern, inspects the JSON schema, and runs a Python aggregation script to compute the reports, forming a discovery-inspection-computation-verification pipeline. The weaker trajectory recovers from an invalid wildcard read by using \texttt{find} and reading the matched files, but does not turn this recovery into a reliable computation process. Consequently, its outputs remain brittle in summary fields, group counts, filtering semantics, and critical-instance checks. This case shows that tool-use appropriateness goes beyond selecting valid tools: reliable agents must compose exploration, computation, feedback, and verification toward the final objective.

\begin{figure}[!t]
\centering
\begin{AIbox}{Case 3: Fine-Grained Instruction Following}

\noindent
{\bf Task requirement.}
{\scriptsize
Generate a reorder plan from \texttt{input/inventory.csv} and
\texttt{input/bulk\_discounts.csv}. The reorder list must include only rows where
\texttt{Quantity <= ReorderPoint}, compute \texttt{TargetStock}, \texttt{OrderQty},
and \texttt{LineTotal} exactly, aggregate supplier orders, and create one JSON file
per supplier under \texttt{output/orders/}.
}

\vspace{0.6em}

\begin{minipage}[t]{0.49\linewidth}
{\bf Strong Model: Rule-Consistent Generation}

\vspace{0.25em}
{\scriptsize
The stronger trajectory computes the outputs from the input tables and verifies the
generated artifacts. It filters rows by the stated reorder condition, aggregates
supplier totals, and creates the complete supplier-level JSON set.
}

\vspace{0.35em}
{\scriptsize\ttfamily
read input/inventory.csv\par
read input/bulk\_discounts.csv\par
\vspace{0.2em}
\goodhl{exec: compute reorder plan from CSVs}\par
\goodhl{\quad keep only Quantity <= ReorderPoint}\par
\goodhl{\quad apply supplier discount rules}\par
\vspace{0.2em}
write output/reorder\_list.csv\par
write output/supplier\_orders.csv\par
\goodhl{write output/orders/*.json}\par
\vspace{0.2em}
orders files:\par
\quad Acme\_Office.json\par
\quad CleanWorks.json\par
\quad Global\_Stationery.json\par
\quad Paper\_Plus\_Co.json\par
\textbf{reward = 1.000}\par
}
\end{minipage}
\hfill
\begin{minipage}[t]{0.49\linewidth}
{\bf Weak Model: Plausible but Flawed Output}

\vspace{0.25em}
{\scriptsize
The weaker trajectory writes structured CSV and JSON files, so the result appears
reasonable at a glance. However, it fails to follow the core filtering rule:
\texttt{output/reorder\_list.csv} should contain only items whose current quantity is
not above the reorder point, but several included rows violate this condition.
}

\vspace{0.35em}
{\scriptsize\ttfamily
write output/reorder\_list.csv\par
...\par
\badhl{\textbf{Required rule:}}\par
\badhl{\quad include row only if}\par
\badhl{\quad Quantity <= ReorderPoint}\par
\vspace{0.2em}
\badhl{\textbf{Actual output violates rule:}}\par
\badhl{\quad Presentation Folders: 12 > 10}\par
\badhl{\quad Correction Tape: 9 > 7}\par
\badhl{\quad Disinfectant Wipes: 8 > 5}\par
\vspace{0.2em}
\badhl{\textbf{Why this is wrong:}}\par
\badhl{\quad these items already exceed their}\par
\badhl{\quad reorder points, so they should not}\par
\badhl{\quad appear in the reorder list}\par
\vspace{0.2em}
write output/supplier\_orders.csv\par
\badhl{\textbf{Propagated error:}}\par
\badhl{\quad supplier totals include invalid rows}\par
\vspace{0.2em}
write output/orders/*.json\par
\badhl{\textbf{Propagated error:}}\par
\badhl{\quad per-supplier order files inherit}\par
\badhl{\quad the invalid reorder items}\par
\textbf{reward = 0.429}\par
}
\end{minipage}



\vspace{0.3em}
\end{AIbox}

\caption{A representative case of fine-grained requirement satisfaction. The weaker trajectory produces plausible reorder artifacts but violates the required \texttt{Quantity <= ReorderPoint} filtering rule, causing invalid rows to propagate into downstream supplier outputs.}
\label{fig:case_fine_grained_requirement_satisfaction}
\end{figure}

\paragraph{Long-Horizon Execution Robustness.}
Long-horizon Claw tasks require agents to coordinate many dependent steps, including reading files, running scripts, generating artifacts, checking intermediate outputs, and preserving workspace state across reruns. In such settings, robustness is not defined by whether every action is correct from the beginning, but by whether the agent can interpret feedback, recover from disruptions, and continue along a valid solution path without losing task context. This is important because early failures often affect later artifacts and become visible in the final workspace state. Figure~\ref{fig:case_error_recovery_long_horizon} illustrates this behavior in a support-ticket automation task. GPT-5.4 encounters a missing optional memory file, treats it as non-blocking, and proceeds to inspect the inbox files. When a cleanup command is rejected by the execution guard, it avoids repeating the failing action, resets the relevant state file, reruns the processor, and verifies that the processed-batch state contains the two expected batches without duplication. In contrast, 30A3 accumulates unresolved errors and reaches an approval-related dead end, leaving the required rewrites, metrics summaries, notifications, and idempotent state incomplete. This case shows that long-horizon robustness depends on maintaining coherent progress under execution feedback, so that local disruptions are corrected before they compound into incomplete task execution.

\paragraph{Fine-Grained Instruction Following.}
Claw-style tasks often contain small but consequential constraints, including filtering rules, output schemas, numeric formulas, and cross-file consistency requirements. Because agents must produce persistent artifacts such as CSV, JSON, Markdown, or scripts, merely creating the requested files does not guarantee task correctness; the artifacts must also satisfy the detailed conditions implied by the instruction and input resources. Figure~\ref{fig:case_fine_grained_requirement_satisfaction} shows this issue in a reorder-plan task. GPT-5.4 first applies the required condition \texttt{Quantity <= ReorderPoint} and then computes downstream supplier outputs from the filtered records. In contrast, 30A3 generates plausible CSV and JSON files but includes invalid items whose quantities exceed their reorder points. These filtering errors propagate into supplier totals and order files, causing the final workspace to appear complete while violating core requirements. This case shows that reliable Claw agents must preserve detailed constraints across generated artifacts and derived outputs, motivating finer-grained supervision over intermediate decisions, artifact contents, and cross-file consistency.

\section{Conclusion}


In this paper, we present ClawGym, a scalable framework that streamlines the end-to-end development pipeline for personal agents by integrating task synthesis, trajectory collection, agent training, and evaluation within Claw-style environments. The framework first introduces a twofold synthesis approach that combines top-to-bottom topic-driven generation with bottom-to-top skill composition, supported by automated environment construction and hybrid verification mechanisms. This synthesis stage produces 13.5K task samples, forming ClawGym-SynData. Building on these synthesized tasks, we then collect 24.5K interaction trajectories on the OpenClaw harness to train ClawGym-Agents. Additionally, we explore the potential of RL through a lightweight sandbox-parallel pipeline. For rigorous evaluation, we further establish ClawGym-Bench, a human-verified benchmark consisting of 200 high-quality instances. Our experiments demonstrate that SFT on synthesized data significantly improves performance across model scales. Specifically, Qwen3-8B achieves gains of 38.90\% on PinchBench and 43.46\% on ClawGym-Bench, while Qwen3-30B-A3B shows improvements of 54.68\% and 25.96\%, respectively.
Furthermore, our behavioral analyses characterize how Claw agents act in environment-grounded tasks, revealing key limitations and highlighting directions for future research on Claw agent development.



\bibliography{ref}
\newpage
\appendix
\section{Evaluation Prompt}\label{app:prompt}

\begin{PromptBox}[p:judge_prompt]{Rubric-Based Judge}
\begin{lstlisting}
# System prompt:

You are a strict rubric-based evaluator. Use only the user prompt content. Do not call tools, browse, inspect files, or ask for more context. Your response may include concise analysis, but it must end with exactly one standalone JSON object with keys `scores` and `notes`.


# User prompt:

You are grading an OpenClaw agent result.

You must not call, request, or simulate any tools. Do not browse, list directories, open files, inspect the workspace, or ask for more context. Grade only from the task, final output files, optional transcript evidence, and rubrics included in this prompt.

Before giving the final judgement, provide concise analysis explaining how the final outputs satisfy or fail each rubric criterion.

Your final judgement must end with exactly one standalone JSON object and nothing else after it. The final JSON object must contain exactly two keys: `scores` and `notes`.

- `scores` must map every rubric id (`criterion_1`, `criterion_2`, ...) to one numeric score chosen from that rubric's allowed score anchors.
- `notes` must be a concise string summarizing the main reasons for the assigned scores.

Do not include any overall score in the final JSON. Do not compute or report the final aggregated score. Score aggregation will be handled separately by post-processing.

## User Task
{{USER_TASK}}

## Final Output Files
{{FINAL_OUTPUT_FILES}}

## Additional Changed Workspace Files
{{ADDITIONAL_CHANGED_WORKSPACE_FILES}}

## Transcript
{{TRANSCRIPT_OPTIONAL}}

## Rubric
{{RUBRIC}}

\end{lstlisting}
\end{PromptBox}



\end{document}